\RequirePackage{fix-cm}
\documentclass[smallcondensed]{svjour3}     
\smartqed  
\usepackage{xcolor}
\usepackage{comment}
\usepackage{caption}
\usepackage{graphicx}
\usepackage{url}
\usepackage{amsmath}
\usepackage{amssymb}
\usepackage{bm}
\usepackage{multirow}
\usepackage{amsfonts}
\usepackage{enumitem}
\usepackage{booktabs}
\usepackage[numbers]{natbib}
\usepackage{changes}
\usepackage{todonotes}

\begin{document}

\title{Temporal Stability in Predictive Process Monitoring
}


\author{Irene Teinemaa \and Marlon Dumas \and Anna Leontjeva \and Fabrizio Maria Maggi}


\institute{University of Tartu, Juhan Liivi 2, 50409 Tartu, Estonia,\\ \email{\{irene.teinemaa,marlon.dumas,anna.leontjeva,f.m.maggi\}@ut.ee}}

\date{Received: date / Accepted: date}

\maketitle

\begin{abstract}
Predictive process monitoring is concerned with the analysis of events produced during the execution of a business process in order to predict as early as possible the final outcome of an ongoing case. Traditionally, predictive process monitoring methods are optimized with respect to accuracy. However, in environments where users make decisions and take actions in response to the predictions they receive, it is equally important to optimize the stability of the successive predictions made for each case. To this end, this paper defines a notion of temporal stability for binary classification tasks in predictive process monitoring and evaluates existing methods with respect to both temporal stability and accuracy. We find that methods based on XGBoost and LSTM neural networks exhibit the highest temporal stability. We then show that temporal stability can be enhanced by hyperparameter-optimizing random forests and XGBoost classifiers with respect to inter-run stability. Finally, we show that time series smoothing techniques can further enhance temporal stability at the expense of slightly lower accuracy.
\end{abstract}

\section{Introduction}
\label{sec:introduction}

Modern organizations generally execute their business processes on top of process-aware information systems, such as Enterprise Resource Planning (ERP) systems, Customer Relationship Management (CRM) systems, and Business Process Management Systems (BPMS), among others~\cite{FBPM}. These systems record a range of events that occur during the execution of the processes they support, including events signaling the creation and completion of business process instances (herein called \emph{cases}) and the start and completion of activities within each case.

Event records produced by process-aware information systems can be extracted and pre-processed to produce business process \emph{event logs}~\cite{van2016process}. A business process event log consists of a set of \emph{traces}, each trace consisting of the sequence of event records produced by one case.
Each event record consists of a number of attributes. Three of these attributes are present in every event record, namely the \emph{event class} (a.k.a.\ \emph{activity name}) specifying which activity the event refers to, the \emph{timestamp} specifying when did the event occur, and the \emph{case id} indicating which case of the process generated this event. In other words, every event represents the occurrence of an activity at a particular point in time and in the context of a given case. An event record may carry additional attributes.
These attributes may be categorical, numerical, or textual.
For example, in a sales process, an event corresponding to activity \emph{payment} could record the \emph{amount} of the payment, the \emph{type} of payment (e.g., cash or by credit card), and an \emph{error message} containing the type of error in case of a failing credit card transaction.
Some attributes vary from one event to another. These are called \emph{event-specific attributes} (or \emph{event attributes} for short). For example, in a sales process, the amount of the payment is specific of activity payment. Other attributes, namely \emph{case attributes}, belong to the case and are hence shared by all events generated by the same case. For example in a sales process, the customer identifier is likely to be a case attribute. If so, this attribute will appear in every event of every case of the sales process, and it will have the same value for all events generated by the same case.

\emph{Predictive process monitoring}~\cite{maggi2014predictive} is a family of techniques that use event logs to predict how an ongoing case (a \emph{case prefix}) will unfold up to its completion. A predictive process monitoring technique may provide predictions on the remaining execution time of each ongoing case of a process~\cite{Rogge-SoltiW13}, the next activity that will be executed in each case~\cite{evermann2017predicting}, or the final outcome of a case wrt.\ a set of possible outcomes~\cite{MetzgerLISFCDP15,teinemaa2017outcome}.
In this work, we concentrate on the latter type of predictions, namely on \emph{outcome-oriented predictive process monitoring}~\cite{teinemaa2017outcome}, where the outcome is assumed to be a binary value (multi-class outcomes are out of scope of this paper). In this context, the outcome of a case can be defined in different ways, depending on the business goals and targets of the process. For instance, in a sales process a desirable outcome is that the customer places a purchase order, while a negative outcome occurs when the customer terminates the process before placing an order.

A variety of outcome-oriented predictive process monitoring techniques have been proposed in the literature~\cite{teinemaa2017outcome}. In existing work, the quality of these methods is measured in terms of prediction accuracy using, for example, precision, recall, and Area Under the ROC Curve (AUC). 
However, we argue that these accuracy measures are not sufficient to assess a predictive process monitoring method. Consider, for instance, a healthcare process where the target is to estimate whether a patient will need intensive or standard care. An accurate prediction could help the patient to receive the suitable treatment in a timely manner, as well as help the hospital to better allocate resources to patients. Suppose that when the patient first arrives at the hospital, the predictor estimates that she will need intensive care, so she is admitted to the intensive care program. After executing a procedure, the predictor changes the prediction and estimates that standard care is sufficient for the patient, so the patient is brought to standard care. However, after performing another procedure, the classifier changes the prediction again and recommends transferring the patient back to intensive care. This example shows how the practical usability of a predictor is limited if it outputs unstable predictions, i.e., if it tends to often change the value of the predictions after seeing new data about the same case. In this example, the treatment of the patient could have been more efficient if the personnel had not trusted the intermediate prediction of the predictor and had not brought the patient to the standard care. Another example concerns a debt encashment process, where a prediction engine can be used to decide whether the debt should be sent to a credit collection agency or not. In this case, volatile predictions can mislead users of the system to prematurely send the debt to the collection agency, resulting in smaller revenue as compared to waiting some more time for the debt to be repaid. Similarly, in case of fraud detection in a financial institution, unstable predictions may cause the institution to frequently block and unblock the credit of a user, resulting in inconveniences and loss of revenue related to potential transactions that the user was not able to complete.

The above examples illustrate the importance of the stability of a classifier when used to make successive predictions in the context of predictive process monitoring.
The conventional notion of stability in non-deterministic learning algorithms (such as random forest) indicates how much the predictions made for the same ongoing case differ across different runs of training the classifier~\cite{elisseeff2005stability}. In other words, if we train multiple classifiers with the same parameter setting but different randomization parameters, would these classifiers agree on the predictions made for the same sample or not? From hereinafter, we refer to this notion of stability as the \emph{inter-run stability}. Conversely, in this paper, we are interested in another type of question, i.e., on how different are the predictions made by the same classifier (or an ensemble of classifiers) for different prefixes of the same case. Specifically, we want to measure whether the classifier often changes its prediction about the same case when more events in the case are performed. We refer to the latter notion of stability as the \emph{temporal stability}. 

In this paper, we:
\begin{enumerate}
\item introduce a measure of temporal stability for binary classification tasks in predictive process monitoring,
\item perform an evaluation of several existing predictive process monitoring methods with respect to both prediction accuracy and temporal stability,
\item study the effects on temporal stability of increasing inter-run stability in combination with prediction accuracy,
\item study the effect on temporal stability and accuracy of applying smoothing techniques to the time series of predictions made for a given case.
\end{enumerate}

The rest of the paper is structured as follows. Section~\ref{sec:related} summarizes the related work on predictive process monitoring, early sequence classification, and learning algorithm stability. Section~\ref{sec:temporal_stability} defines the notion of temporal stability and proposes a metric for measuring it, as well as a post-processing technique to combine predictions made for prefixes of the same case in order to reduce their volatility. Section~\ref{sec:evaluation} describes the experimental set-up and the results of the evaluation. Section~\ref{sec:conclusion} concludes the paper and discusses avenues for future work.

\section{Related Work}
\label{sec:related}

In this section, we discuss the related work on predictive process monitoring, early sequence classification, and stability in learning algorithms.

\subsection{Predictive Process Monitoring}
\label{sec:related_ppm}
A variety of predictive process monitoring methods have been proposed in the existing literature~\cite{marquez2017predictive}. These approaches can be divided according to the prediction target into the following categories: remaining time prediction (regression tasks), next activity prediction (multi-class classification), and outcome-oriented prediction (binary classification). Outcome-oriented process monitoring techniques differ in terms of three aspects: sequence encoding, bucketing of prefixes (how many classifiers are built and which prefixes are given as input to each classifier), and classification algorithm~\cite{teinemaa2017outcome}.

A sequence encoding can be \emph{lossless}, meaning that the original trace can be recovered completely from the encoded trace. An example of such encoding is the \emph{index-based encoding} proposed by Leontjeva et al.~\cite{leontjeva2015complex}, which concatenates the data from all events into a single vector, so that the first position contains the activity name from the first event, the second position contains the activity name from the second event and so on. A drawback of this method is that the size of the encoded vector increases with each event, which means that a separate classifier is needed for each prefix length. Alternatively, a \emph{lossy} encoding approach aggregates the event data for each trace, thus producing feature vectors of the same size independently of the prefix length. Examples of lossy encodings are \emph{last state encoding}, which uses only data from the most recent event performed in each trace and \emph{aggregation encoding}, which aggregates the information from all events executed so far using, for instance, the frequencies of categorical event attributes (e.g., activity names), or aggregation functions such as minimum, mean, or maximum for numeric event attributes.
Using a lossy encoding, we can feed all the encoded prefixes to a \emph{single classifier}, as the length of the feature vector does not depend on the prefix length.

Several existing works have proposed dividing prefixes into buckets and training separate classifiers for each bucket, resulting in a \emph{multiclassifier} approach. An example is~\cite{leontjeva2015complex}, where different classifiers are built for each prefix length. Other methods cluster the prefixes based on their similarity in terms of the performed activities and build one classifier per cluster~\cite{di2016clustering}. Others train a classifier for every state in a process model or in a transition system~\cite{lakshmanan2010predictive}.

Existing works have experimented with different classification algorithms. The most popular choices are tree-based methods, such as decision trees~\cite{di2016clustering,lakshmanan2010predictive,de2016general} and random forests~\cite{leontjeva2015complex,di2016clustering}.
To our knowledge, there is no existing work on using recurrent neural networks (RNNs) for outcome-oriented predictive process monitoring. However, RNNs with long short term memory units (LSTMs) have been used in other predictive process monitoring tasks, such as for predicting the remaining time and the next activity~\cite{tax2017predictive,evermann2017predicting}.

\subsection{Early Sequence Classification}

With respect to the broader literature on machine learning, outcome-oriented predictive process monitoring is related to \emph{early sequence classification}. Given a set of labeled sequences, the goal is to build a model that for a sequence prefix predicts the label this prefix will get when completed. A survey on sequence classification presented in~\cite{santos2016literature} gives an overview of the techniques from this field.

Xing et al.~\cite{xing2008mining} introduced the notion of \emph{seriality} in sequence classifiers, referring to the property that for each sequence, there exists a prefix length starting from which the classifier outputs (almost) the same prediction. The works on early sequence classification are generally focused on determining such prefix length that yields a good prediction, also referred to as the \emph{minimal prediction length} (MPL)~\cite{xing2012early}. The method by Xing et al.~\cite{xing2012early} finds the earliest timestamp when the nearest neighbor relationships in the training data become stable (i.e., remain the same in the subsequent prefixes).
Parrish et al. proposed a method based on the \emph{reliability} of predictions, i.e., the probability that the label assigned to a given prefix is the same as the label assigned to the whole sequence~\cite{parrish2013classifying}. More recently, Mori et al.~\cite{mori2017reliable} designed an approach to make an early prediction when the ratio of accuracy between the prediction made for the prefix and for the full sequence exceeds a predetermined threshold.
Most of the techniques for early classification are designed for numerical time series or simple (univariate) symbolic sequences. However, the problem of predictive process monitoring can be seen as an early classification over complex sequences where each element has a timestamp, a discrete attribute referring to an activity, and a payload made of a heterogeneous set of other attributes. One of the few works on early classification on complex sequences is \cite{lin2015reliable}, where Lin et al. propose constructing \emph{serial decision trees} and monitor the error rate in leaf nodes in order to determine the MPL.

The works on developing serial classifiers and finding the MPL are closely related to the notion of temporal stability studied in this paper. In fact, a serial classifier has perfect temporal stability. However, instead of determining MPL and making predictions only after the MPL is reached, we are interested in predicting the outcome for every prefix of the sequence. The reason for this is that in a predictive process monitoring setting, it is necessary to give the best estimate of the case outcome even when too few data is available to make a final prediction. In this respect, we aim for temporal stability also on short prefixes, when the prediction might still differ from the one that would be made for the entire sequence.

\subsection{Stability of Learning Algorithms}
Stability of learning algorithms has been a topic of interest for many years. Conventionally, a learning algorithm is considered unstable if small changes (perturbations) in the training set can cause significant changes in the predictor~\cite{breiman1996bagging}. Such instability of single predictors motivated Breiman et al. to introduce \emph{bagging predictors}, showing that the stability and accuracy of a predictor can be increased by aggregating the estimations from multiple versions of the predictor~\cite{breiman1996bagging}. In this context, increasing stability relates to decreasing the variance between prediction estimates.
Bousquet et al. studied the relationship between stability and generalization~\cite{bousquet2002stability}. In particular, their study is based on \emph{sensitivity analysis}, i.e., how much replacing or deleting a training sample affects the prediction loss. They propose three definitions of stability, which are all based on changes in the training set. The reason for this is that they focus on deterministic algorithms, so that all the randomness comes from the sampling on the datasets.
Elisseeff et al. extended these notions of stability to non-deterministic algorithms~\cite{elisseeff2005stability} where randomness is present even when the training set remains unchanged. Their stability definitions are supplemented with a randomness parameter.
More recently, Liu et al. proposed a metric for measuring stability across several runs of random forest and incorporated it into a framework for selecting the hyperparameters based on a goodness measure combining AUC, stability, and cost~\cite{liu2017generalising}.

While existing notions of stability are related to changes made in the training phase (either by changing the training set or by changing the randomness parameter), in this paper we study the case where both the training dataset and the randomness are fixed, but the input vector changes over time.
In particular, we study the \emph{temporal stability} of predictions in the setting where predictions are made successively for different prefixes of the same sequence. In other words, we examine how much increasing the length of the prefix changes the predictions.

\section{Temporal Prediction Stability}
\label{sec:temporal_stability}

In this section, we start with introducing the notion of prediction scores in outcome-oriented predictive process monitoring. We proceed with defining temporal stability and provide a metric to measure this property. Lastly, we describe our approach for combining prediction scores obtained for prefixes of the same case in order to reduce their volatility.

\subsection{Prediction Scores Over Time}
\label{sec:prediction_scores}

In an outcome-oriented predictive process monitoring task, the target for classification is a binary value, referring to either a positive or a negative outcome. Despite the fact that the classifier is trained to recognize a binary target, it can usually output a real-valued \emph{prediction score} indicating the likelihood towards the positive outcome. 

\begin{figure}[hbtp]
\centering
\includegraphics[width=0.9\textwidth]{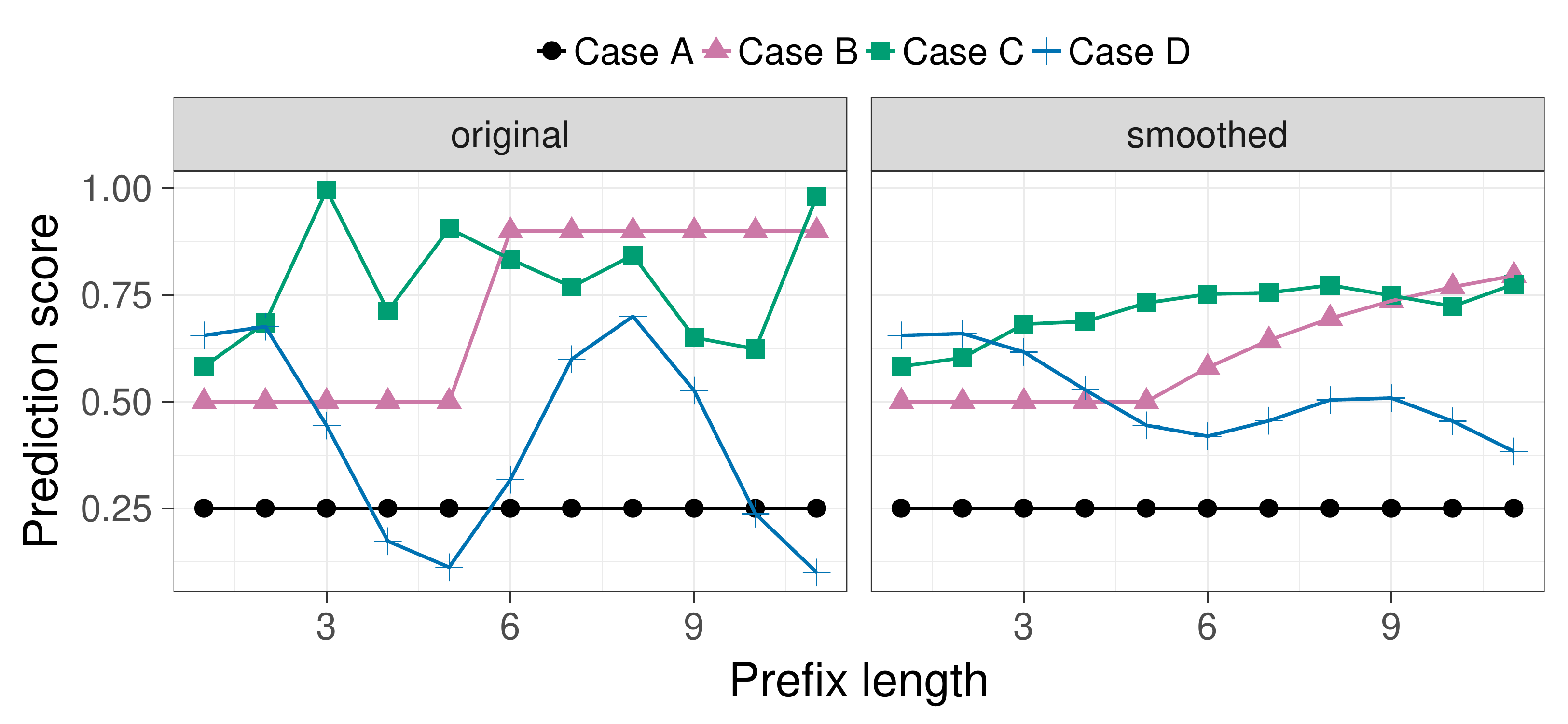}
\caption{Examples of prediction scores over time: original (left) and smoothed (right).}
\label{fig:example_prediction_scores}
\end{figure}

In predictive process monitoring, the classifier is asked to give an estimation about the case outcome after each performed event. Therefore, the prediction scores estimated after each event of the same case form a time series. As an example, consider the pink time series (Case B) plotted in \figurename~\ref{fig:example_prediction_scores} (left). During the first 5 events, the classifier is unsure about what will be the outcome of this case (the prediction scores for these events are equal to 0.5). Then, the 6th event provides some relevant signal, so that the classifier becomes confident that the case will be positive (the prediction scores for the following events are 0.9). This series is rather stable over time, as the successive prediction scores change only once. An example of a completely stable series of predictions is Case A (the black line), where the prediction scores remain the same for all prefixes.
Now consider Cases C and D (green and blue). We can see that the classifier changes the prediction score after almost every event, producing a \emph{volatile} time series for these cases. Such unstable predictions have little practical value, causing users to be cautious about acting upon the prediction and decreasing the overall credibility of the classifier.

\subsection{Temporal Stability}
\label{sec:stability_notion}

Based on the above rationale, we say that a classifier is \emph{temporally stable} if it (generally) outputs similar predictions to successive prefixes from the same sequence.

Given a threshold on the prediction scores that determines whether the predicted outcome is positive or negative, it would be natural to define temporal instability as the number of times the classifier ``flips'' its prediction, and to define temporal stability as one minus a normalized measure of instability. The drawback of this approach is that it is dependent on the chosen threshold. Instead, we aim for a more general, threshold-independent measure that would capture the stability of the classifier under any threshold. Accordingly, we propose to measure stability as a function of the magnitude of the changes between successive prediction scores. This latter definition is related to the former: If the difference between successive scores is high, there exist many thresholds that would lead to flips in the predicted outcome. Conversely, if the difference is low, only a low number of thresholds would flip the prediction.

The simplest way to consider the magnitude of the changes would be to measure the (average) absolute difference between successive prediction scores. Note that this metric does not consider the direction of the changes, i.e., a change towards the correct direction (the actual class) affects the measure in the same way as a change towards the wrong direction. As a result, a classifier that consistently improves its prediction is assigned a similar stability score as one that fluctuates around the same score. An alternative would be to consider only the changes that are made to the wrong direction, calculating the (average) absolute difference only over these changes. However, this metric would reflect the \emph{consistency} of the classifier rather than its \emph{stability}. For instance, consider a sequence with the actual outcome being \emph{positive}, and two classifiers. One of the classifiers outputs a score of 1 at the first event, i.e., it is (correctly) certain that the outcome will be positive, but throughout the case becomes only slightly less certain of it, outputting 0.99 on some events. The other classifier makes a completely wrong estimation at the beginning of the sequence, outputting a score of 0, while throughout the rest of the case, it only improves its estimate (sometimes by large magnitudes), producing scores like 0.1, 0.5, and even 0.95. According to the latter metric, the second classifier, which makes changes in large magnitudes, would be considered more stable than the first classifier, although the first one only changes its prediction by a small amount. In a sense, a measure that considers the direction of the change penalizes classifiers that make the right prediction from the onset, since the only way to maintain their stability throughout the sequence would be to always output exactly the same score. Based on these considerations, we proceed with measuring the average difference between the successive prediction scores without taking into account the direction of the change.

Accordingly, we measure the temporal stability (TS) of a classifier as one minus the average absolute difference between any two successive prediction scores:

\begin{equation}
\label{eq:temporal_stability}
TS = 1 - \frac{1}{n} \sum_{i=1}^{n} \frac{1}{T_i-1} \sum_{t=2}^{T_i} |\hat{y}_t^i-\hat{y}_{t-1}^i|,
\end{equation}
where $n$ is the number of cases used for the evaluation, $T_i$ is the total number of events in the $i$-th case, and $\hat{y}_t^i$ is the prediction score of the $t$-th event of the $i$-th case.
This metric first evaluates the average absolute difference between successive prediction scores within each case in order to eliminate the bias towards long sequences, and then averages over the cases.

\subsection{Combining Prediction Scores via Smoothing}

We can adjust the prediction scores during a post-processing phase to reduce volatility without affecting the pre-trained classifier. Specifically, instead of using explicitly the score that the classifier outputs for a case after observing $t$ events, we combine it with prediction scores made for shorter prefixes of the same case.

To combine predictions, we can use various time series \emph{smoothing} methods, which average out the noise and fluctuations.
The simplest way to smooth a time series is via a \emph{moving average}. The smoothed estimate at each event is computed as the average of the last $k$ observations. A different approach, called \emph{single exponential smoothing}, assigns weights that decrease exponentially over time. The smoothed estimate at time $t$ is the combination of the observed value at time $t$ and the smoothed estimate at time $t-1$, using a smoothing parameter $\alpha$, $0 <= \alpha <= 1$: $s_t = (1 - \alpha) \cdot \hat{y}_{t} + \alpha \cdot s_{t-1}$. Parameter $\alpha$ controls to what extent the previous observations are taken into account. The larger the $\alpha$, the stronger the smoothing effect. While other smoothing techniques are available, we use the single exponential smoothing because of its simplicity and because it allows us to directly control the level of smoothing.
Also, only techniques that enable sequential smoothing (as opposed to smoothing over the entire sequence) are applicable in our case, as in the predictive process monitoring setting, only the prediction scores made up to a certain point in the sequence are known.

For example, consider the time series plotted in \figurename~\ref{fig:example_prediction_scores} (right). These time series have been derived from the examples in \figurename~\ref{fig:example_prediction_scores} (left) by applying exponential smoothing with $\alpha = 0.8$. We can notice that the fluctuations in Cases C and D have been reduced considerably.
However, smoothing can also have a negative effect on the predictions, illustrated by Case B. Namely, changes in the scores do not have an immediate strong effect, as the adjusted score puts some weight on the previous estimates. Therefore, when an event carrying a relevant signal about the case outcome arrives, the smoothed estimate is cautious about trusting it, resulting in a lower accuracy.

\section{Evaluation}
\label{sec:evaluation}

We conducted an empirical evaluation to address the following questions:

\begin{enumerate}[label=RQ\arabic*,leftmargin=*]
\item What is the relative performance of different predictive process monitoring methods in terms of temporal stability (in addition to accuracy)?
\item How does maximizing the inter-run stability in combination with prediction accuracy affect the temporal stability?
\item How does decreasing prediction volatility via exponential smoothing affect the accuracy and the temporal stability?
\end{enumerate}

Below, we describe the approaches and datasets employed, we explain the experimental setup, and discuss the results. The code used for this evaluation is available at \url{https://github.com/irhete/stability-predictive-monitoring}.

\subsection{Approaches}

To address RQ1, we choose 7 predictive process monitoring approaches (see Table \ref{table:approaches}) as basis for the experiments. We employ 2 existing sequence encoding techniques, the index-based and the aggregation encoding. As explained in Section~\ref{sec:related_ppm}, index-based encoding constructs a lossless representation of a prefix by concatenating the data from each executed event. In the aggregation encoding, a prefix of arbitrary length is transformed into a fixed length feature vector by applying different aggregation functions. In particular, for categorical features, we use frequencies, i.e., how many times each possible value (e.g., each activity name) has occurred in the given prefix, while the numerical features are aggregated using the average, maximum, minimum, sum, and standard deviation of the values observed so far.
Both encodings are combined with two classification methods, random forest~\cite{breiman2001random} (RF) and XGBoost~\cite{chen2016xgboost}. We choose these classifiers because they have shown to outperform other methods in various applications~\cite{fernandez2014we, olson2017data}. Additionally, we adapt a predictive process monitoring method based on LSTM neural networks~\cite{tax2017predictive} to predict the outcome of a case.

In all of the approaches, each prefix constitutes a separate training instance. 
For index-based encoding, the fact that different prefixes consist of different numbers of events raises an issue when trying to encode all prefixes with fixed-length vectors.
There are two possible solutions to this issue. Firstly, it is possible to fix the maximum prefix length and, for shorter prefixes, pad the data for missing events with zeros. An alternative solution is to build multiple classifiers, one for each prefix length; given a prefix of length $l$ in the testing set, the prediction for this prefix is derived from the classifier constructed based on prefixes (in the training set) of length $l$.
In our experiments, we apply both solutions to the RF and XGBoost based approaches, marked as \emph{RF\_idx\_pad}/\emph{XGB\_idx\_pad} and \emph{RF\_idx\_mul}/\emph{XGB\_idx\_mul}, respectively. Since the second, multiclassifier solution is not commonly used with LSTMs, in the LSTM-based approach we only apply the padding solution.

Prediction scores returned by classifiers are often poorly calibrated, meaning that the scores do not reflect well the actual probabilities of belonging to one class or to the other~\cite{guo2017calibration}. For instance, one classifier may output scores that are always concentrated around 0.5, while another may return scores that are well distributed within the range between 0 and 1. This causes bias when comparing different classifiers in terms of temporal stability. Indeed, the differences between any two prediction scores in the case of the former classifier are very small, making it seem a very stable classifier, while the relative differences within each case might be larger than in the latter classifier.
To address this issue, we apply a well-known calibration method, Platt scaling~\cite{platt1999probabilistic}, to each of the classifiers before comparison. We choose this technique because it outperforms other methods when data is scarce (e.g., less than 1000 data points available for calibration)~\cite{niculescu2005predicting}, which is the case in most of our datasets.
Note that calibration does not change the order of the prediction scores assigned by the same classifier, so that the AUC of each classifier is not affected by it.

To test RQ2, we adapt the approach proposed in~\cite{liu2017generalising} to RF and XGBoost hyperparameter optimization. Namely, instead of choosing the optimal parameter setting based on AUC on a single run of classifier training, we perform 5 runs with each setting and choose the one that achieves 1) the best average AUC over all runs, and 2) the best combined AUC and inter-run stability\footnote{\emph{Inter-run stability refers to the MSPD metric introduced in~\cite{liu2017generalising}: $MSPD(f) = 2\mathbb{E}_{x_i}[Var(f(x_i)) - Cov(f_j (x_i), f_k(x_i))],$ where $\mathbb{E}_{x_i}$ is the expectation over all validation data, $f$ is a mapping from a
sample $x_i$ to a label $y_i$ on a given run, $Var(f(x_i))$ is the variance of the predictions of a single data point over the model runs, and $Cov(f_j (x_i), f_k(x_i))$ is the covariance of predictions of a single data point over two model runs.}} over all runs. For the latter scenario, we give more weight to the inter-run stability, assigning weights 1 and 5 to AUC and stability, respectively.

To decrease prediction volatility (RQ3), we experiment with exponential smoothing, varying the smoothing parameter $\alpha \in \{0.1, 0.25, 0.5, 0.75, 0.9\}$.

\begin{table}[t]
\caption{Approaches.}
\label{table:approaches}
\begin{center}
{\scriptsize
\begin{tabular}{@{}lccc@{}}
\toprule
Approach & Multi/single cls & Encoding & Classifier \\
\midrule
RF\_agg & single & aggregation & RF \\
RF\_idx\_pad & single & index & RF \\
RF\_idx\_mul & multi & index & RF \\
XGB\_idx\_pad & single & index & XGBoost \\
XGB\_idx\_mul & multi & index & XGBoost \\
XGB\_agg & single & aggregation & XGBoost \\
LSTM & single & index & LSTM \\
\bottomrule
\end{tabular}}
\end{center}
\end{table}

\subsection{Datasets}

We use real-life datasets publicly available at the 4TU Centre for Research Data~\footnote{Production log: \\ \url{https://data.4tu.nl/repository/uuid:68726926-5ac5-4fab-b873-ee76ea412399}, \\ other logs: \url{https://data.4tu.nl/repository/collection:event_logs_real}}.
From the 4TU Centre datasets, we left out those that are not business process event logs, but instead related to software development or web services. Moreover, we excluded event logs where a natural labeling for the case outcome was not easily derivable. Also, we discarded the datasets where the order of events is not clearly defined due to time granularity issues.
For each selected log, it is possible to come up with multiple definitions of case outcome (\emph{labelings}), so that each definition constitutes a separate predictive process monitoring problem. In the following, we briefly describe the domain of each of the datasets and the labelings that were constructed for carrying out the experiments. Then, we describe the feature extraction and preprocessing principles applied to the datasets and conclude with a comparison of general statistics of the datasets.

\paragraph{BPIC2012.} This dataset, originally published in relation to the Business Process Intelligence Challenge (BPIC) in 2012, contains the execution history of a loan application process in a Dutch financial institute. Each case in this log records the events related to a particular loan application. For classification purposes, we defined some labelings based on the final outcome of a case, i.e., whether the application is accepted, rejected, or cancelled. Intuitively, this could be thought of as a multi-class classification problem. However, to remain consistent with previous work on outcome-oriented predictive process monitoring, we approach it as three separate binary classification tasks. In the experiments, these tasks are referred to as \emph{bpic2012\_accepted}, \emph{bpic2012\_declined}, and \emph{bpic2012\_cancelled}.

\paragraph{BPIC2017.} This event log originates from the same financial institution as the BPIC2012 one. However, the data collection has been improved, resulting in a richer and cleaner dataset. As in the previous case, the event log records execution traces of a loan application process. Similarly to BPIC2012, we define three separate labelings based on the outcome of the application, referred to as \emph{bpic2017\_accepted}, \emph{bpic2017\_refused}, and \emph{bpic2017\_cancelled}.

\paragraph{Sepsis cases.} This log records trajectories of patients with symptoms of the life-threatening sepsis condition in a Dutch hospital. Each case logs events since the patient's registration in the emergency room until her discharge from the hospital. Among others, laboratory tests together with their results are recorded as events. Moreover, the reason of the discharge is available in the data in an obfuscated format.

We created three different labelings for this log:

\begin{itemize}
\item \emph{sepsis\_cases\_1}: the patient returns to the emergency room within 28 days from the discharge,
\item \emph{sepsis\_cases\_2}: the patient is (eventually) admitted to intensive care,
\item \emph{sepsis\_cases\_3}: the patient is discharged from the hospital on the basis of something other than \emph{Release A}, which is the most common release type.
\end{itemize}

\paragraph{Hospital billing.} This dataset comes from an ERP system of a hospital. Each case is an execution of a billing procedure for medical services. We created a labeling based on whether the case is reopened or not.

\paragraph{Road traffic fines.} This log comes from an Italian local police force. The dataset contains events about notifications sent about a fine, as well as (partial) repayments. Additional information related to the case and to the individual events include, for instance, the reason, the total amount, and the amount of repayments for each fine. We created the labeling (\emph{traffic\_fines}) based on whether the fine is repaid in full or is sent for credit collection.

\paragraph{Production log.} This log contains data from a manufacturing process. Each trace records information about the activities, workers and/or machines involved in producing an item. The labeling (\emph{production}) is based on whether or not
the number of rejected work orders is larger than zero.

Before encoding the traces for classification, we apply some preprocessing on the raw datasets\footnote{Preprocessed data: \url{https://github.com/irhete/stability-predictive-monitoring}}. In general, we use all the available case and event attributes without doing any feature extraction before encoding. Still, a few extra features are added to each event based on the timestamps, namely, \emph{hour, weekday, month, time since case start}, and \emph{time since last event}. Additionally, we include the \emph{event number}, i.e., how many events have been performed in the case up to the current event. While all these features are calculated \emph{intra-case}, i.e., considering only data from the given case, features could also be extracted \emph{inter-case}, i.e., based on all cases that were active at the time the event was performed. Accordingly, we extract the \emph{number of open cases} (how many cases were open during the execution of the event) as another feature.
Different strategies for extracting inter-case features are discussed in~\cite{senderovich2017intra}.

Each categorical attribute has a fixed number of possible values, called \emph{levels}. For some attributes, the number of distinct levels can be very large, with some of the levels appearing only in a few cases. In order to avoid the dimensionality explosion of the input dataset, we set the category levels that appear in 10 or less samples to a common level \emph{other}.

Due to the fact that event logs consist of data that are recorded automatically by information systems during the execution of tasks of a process, there is none or very little \emph{missing data} in the traditional sense. However, it is common that different events carry different data payloads, resulting in a situation where some attribute values for a given event can be ``missing'' due to the fact that they are not applicable for that particular event. This can be caused by mainly two reasons. Firstly, in most event logs, an event records only the values of data attributes that were changed during that particular event. Therefore, in order to determine the value of an attribute at the point where an event occurred, we need to search for the latest event in the trace (or trace prefix) where the value of the attribute in question changed (or the first event if no change point is found). For instance, the name of the resource involved in the execution of an activity in a case is often logged only if the resource has changed since the previous event. In such cases, we search for the closest preceding event in the same case where the resource name was present and use the same value in the feature vector produced for the current event.
Secondly, different activities can produce different types of data. For instance, in a loan application process, information about the offer made to the customer becomes available only when an offer is made (before that, no offer nor information about it exists). Similarly, in a fine collection process, the amount of payment is only available for payment events. These examples constitute a form of \emph{legitimately missing data}~\cite{osborne2013dealing} or \emph{missing data that is out of scope}~\cite{schafer2002missing}. In our experiments, we decided to address such cases by adding an additional feature (for each data attribute) to the dataset, indicating whether the given value is present for a given event or not. The value of the attribute itself was set to 0 if not present.

In event logs where information is available about case completion, we filter out incomplete cases in order to not mislead the classifier. Also, we cut each trace before the event that was used to define the label. For instance, in the \emph{production} log, the traces are cut immediately before the number of rejected work orders becomes larger than zero.

The datasets (after preprocessing) exhibit different characteristics presented in Table \ref{table:dataset_stats}. Firstly, the number of cases varies from 220 in the \emph{production} log to 129\,615 in the \emph{traffic\_fines} log. Class imbalance is the most severe in the \emph{hospital\_billing} dataset, where only about 5\% of cases are of the positive class. On the other hand, the classes are almost perfectly balanced in the \emph{production}, \emph{traffic\_fines}, \emph{bpic2017\_cancelled}, and \emph{bpic2012\_accepted} datasets. The median trace length is the smallest in \emph{traffic\_fines}, where half of the cases consist of 4 or less events, while BPIC2012 and BPIC2017 variants have the longest traces (median length 35). Trace lengths can be very heterogenous. For instance, while the median trace length in \emph{hospital\_billing} is 6, the maximum trace length is 217. Our experiments have shown that using the original length for very long traces causes the performance of the classifier to decrease, as well as hinders the readability of the plots (see Figures~\ref{fig:aucs_calibrated_not_truncated} and~\ref{fig:stability_mean_abs_calibrated_not_truncated} in the Appendix). Therefore, we have decided to use truncated versions of long sequences. We determined the \emph{truncated length} independently for each dataset based on the following criteria. Firstly, the sequence was truncated from the length where 90\% of the minority class sequences have already completed (and not available anymore for training and evaluation), as both training and evaluation of the classifier would be unreliable when having very few sequences from one of the classes. Secondly, as in the BPIC2012 and BPIC2017 variants the signal starts to converge around 40 and 20 events, respectively, we further truncated the sequences to these lengths for computational reasons. For histograms of case lengths in both classes, see \figurename~\ref{fig:case_length_hist} in Appendix.

\begin{table}[t]
\caption{Dataset statistics.}
\label{table:dataset_stats}
\begin{center}
\resizebox{1\textwidth}{!}{
	\begin{tabular}{@{}lccccccccc@{}}
	\toprule  &  & pos class  & med  & max  & trunc.  &  \\
	dataset name & \# traces &  ratio &  length &  length & length & \# events \\
	 \midrule
	bpic2012\_accepted & 4\,685 & 0.48 & 35 & 175 & 40 & 155\,783 \\
	bpic2012\_declined & 4\,685 & 0.17 & 35 & 175 & 40 & 155\,783 \\
	bpic2012\_cancelled & 4\,685 & 0.35 & 35 & 175 & 40 & 155\,783 \\
	bpic2017\_accepted & 31\,413 & 0.41 & 35 & 180 & 20 & 624\,352 \\
	bpic2017\_refused & 31\,413 & 0.12 & 35 & 180 & 20 & 624\,352 \\
	bpic2017\_cancelled & 31\,413 & 0.47 & 35 & 180 & 20 & 624\,352 \\
	sepsis\_cases\_1 & 782 & 0.14 & 14 & 185 & 29 & 12\,189 \\
	sepsis\_cases\_2 & 782 & 0.14 & 13 & 60 & 13 & 9\,178 \\
	sepsis\_cases\_3 & 782 & 0.14 & 13 & 185 & 22 & 11\,056 \\
	hospital\_billing & 77\,525 & 0.05 & 6 & 217 & 8 & 404\,721 \\
	traffic\_fines & 129\,615 & 0.46 & 4 & 20 & 10 & 460\,462 \\
	production & 220 & 0.53 & 9 & 78 & 23 & 2\,275 \\
	\bottomrule
	\end{tabular}}
\end{center}
\end{table}

\subsection{Experimental Setup}

We apply a temporal split for dividing cases into training and test sets. Namely, the cases are ordered according to the start time and the first 80\% is used for training and validating the models, while the remaining 20\% is used to evaluate the performance. Note that, using this approach, some events in the training cases might still overlap with the test period. As we are using an inter-case feature (the number of open cases), which considers data from all cases active at a given time, this could introduce a bias into our evaluation. In order to avoid that, we cut the training cases so that events that overlap with the test period are discarded.

To achieve the best performance with each method, the hyperparameters of the classifiers need to be optimized separately for each method and dataset. To this end, we further split the training cases randomly into 80\% training and 20\% validation data. We train the models with different parameter settings on the training set and select the model that performed best on the validation set. In the case of RF and XGBoost, the best models are selected based on the AUC on the validation data. During training, LSTMs optimize binary crossentropy, which is why we selected the best parameters according to this metric.

While RF tends to perform well even with little optimization, XGBoost and
LSTM are much more sensitive to hyperparameter selection. Also, the number of
hyperparameters is larger on the last two methods, making grid search infeasible.
In order to keep the methods comparable, we decided to use the same optimization procedure for all of them, i.e., random search~\cite{bergstra2012random} with 16 iterations.
As a basis for random search, we specified for each hyperparameter a distribution to sample values, as well as the bounds for the values (see Table~\ref{table:hyperparameters} in Appendix). The selected values for each hyperparameter are presented in Tables~\ref{table:optimal_params_rf}--\ref{table:optimal_params_interrun} in Appendix. The activation function for LSTM is always fixed to \emph{sigmoid} in our experiments and the number of epochs to 50.

\subsection{Results}

The experiments were performed using Python libraries Scikit-Learn\footnote{\url{http://scikit-learn.org/}} (RF and XGBoost) and Keras\footnote{\url{https://github.com/fchollet/keras/}} with Theano\footnote{\url{http://www.deeplearning.net/software/theano/}} backend (LSTM).

\subsubsection{General comparison}

\begin{figure}[t]
\centering
\includegraphics[width=1\textwidth]{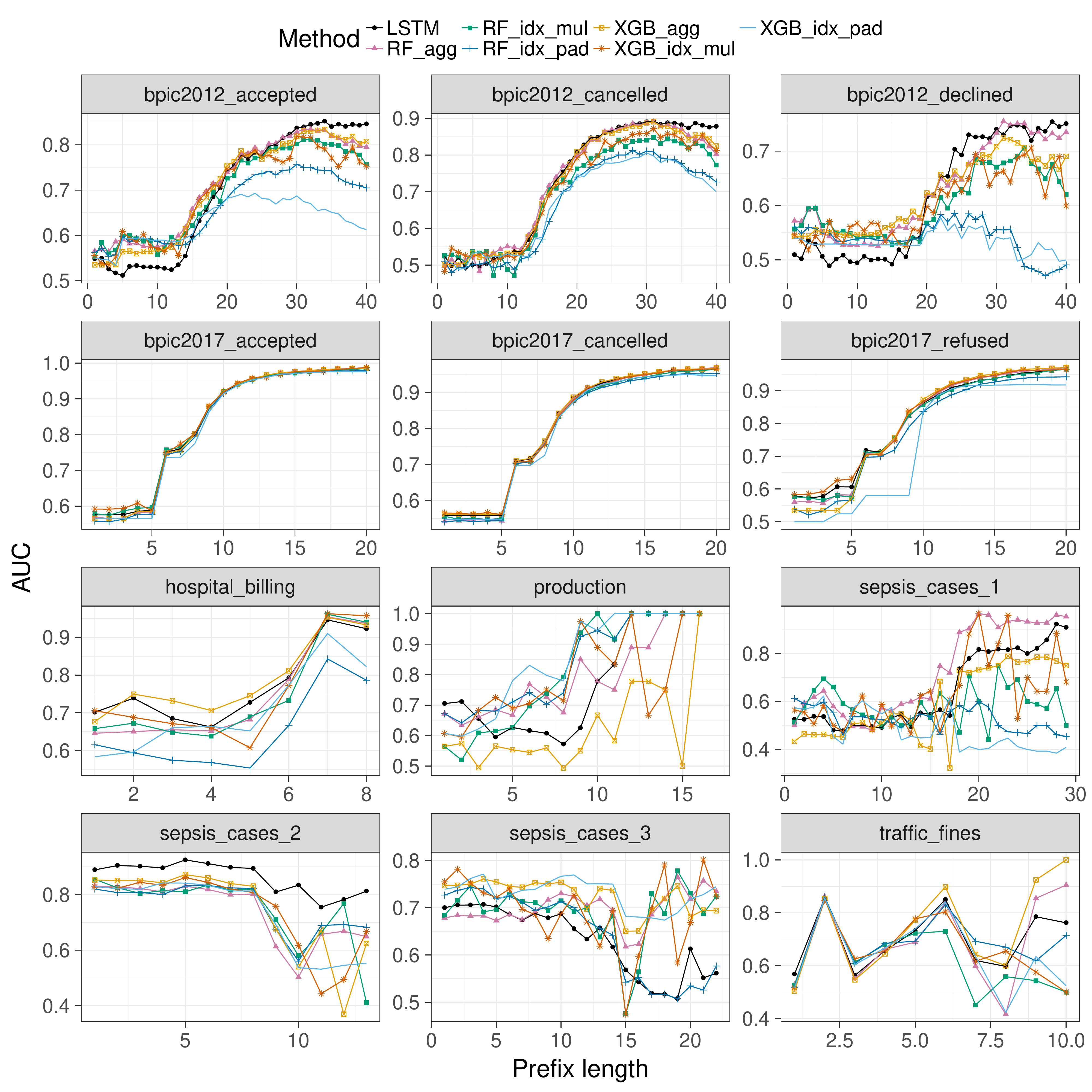}
\caption{Prediction accuracy (measured in terms of AUC).}
\label{fig:aucs_calibrated}
\end{figure}

Figure~\ref{fig:aucs_calibrated} shows the prediction accuracy (AUC) across different prefix lengths. For instance, prefix length 10 means that the predictions were made based on the first 10 events in a case. One observation is that the multiclassifiers (\emph{RF\_idx\_mul} and \emph{XGB\_idx\_mul}) can yield a high accuracy on some prefixes (especially on the shorter ones), but at the same time the results are very volatile, causing the AUC to drop unexpectedly. For instance, see \emph{XGB\_idx\_mul} with $prefix=24$ in \emph{sepsis\_cases\_1} or \emph{RF\_idx\_mul} with $prefix=15$ in \emph{sepsis\_cases\_3}. On long prefixes, the index-based encoding approaches (both multiclassifiers and single classifiers with padding) tend to perform worse than the other methods. Exceptions are some smaller datasets, namely, \emph{production} and \emph{sepsis\_cases\_3}, where \emph{XGB\_idx\_pad} performs well over all prefix lengths.

Different patterns can be seen for \emph{LSTM}. Firstly, in the case of \emph{bpic2012} variants, the accuracy is lower for shorter prefixes, but after the relevant signal comes in (around prefix length between 12 to 20), the model is able to make use of it better than the other methods, reaching the highest AUC on long prefixes. Secondly, while \emph{LSTM} often does not achieve the highest AUC, it is always reasonably stable, in the sense that no sudden drops in AUC occur in any prefix length.

The single classifiers with aggregation encoding (\emph{RF\_agg} and \emph{XGB\_agg}) perform well on both short and long prefixes. Although in some prefix lengths they are outperformed by the index-based encoding methods, they are overall more stable. In particular, these methods are somewhat more volatile than \emph{LSTM}, but they usually do not undergo strong falls in AUC as the multiclassifiers. For example, see \emph{sepsis\_cases\_1} and \emph{sepsis\_cases\_3} where \emph{RF\_agg} and \emph{XGB\_agg} retain high accuracy on long prefixes, while \emph{RF\_idx\_mul} and \emph{XGB\_idx\_mul} become more volatile.

We can also observe, in \figurename~\ref{fig:aucs_calibrated}, that, in some cases, the AUC starts to decline as the prefix length increases, which is counter-intuitive since the longer the prefix, the more information the classifier has to make a decision. For instance, this happens in the \emph{bpic2012} variants, \emph{sepsis\_cases\_2}, \emph{sepsis\_cases\_3}, and \emph{traffic\_fines}. To investigate this phenomenon, we filtered out the short cases, leaving only those that reach the maximum considered prefix length, and calculated the AUC only for those long cases. We observed (\figurename~\ref{fig:aucs_calibrated_max_prefix} in Appendix) that the AUC does not undergo a decrease when considering only the long cases, but instead keeps increasing (or stays at the same level). These results suggest that the decrease in AUC is not due to the classifiers starting to perform worse on long prefixes. Rather, this decrease is due to the fact that for shorter cases, it is easier to make predictions since they are initially closer to completion. Therefore, after these cases have completed and they are excluded from the calculation of the AUC, the performance of the classifier seems to decay.

\begin{figure}[t]
\centering
\includegraphics[width=1\textwidth]{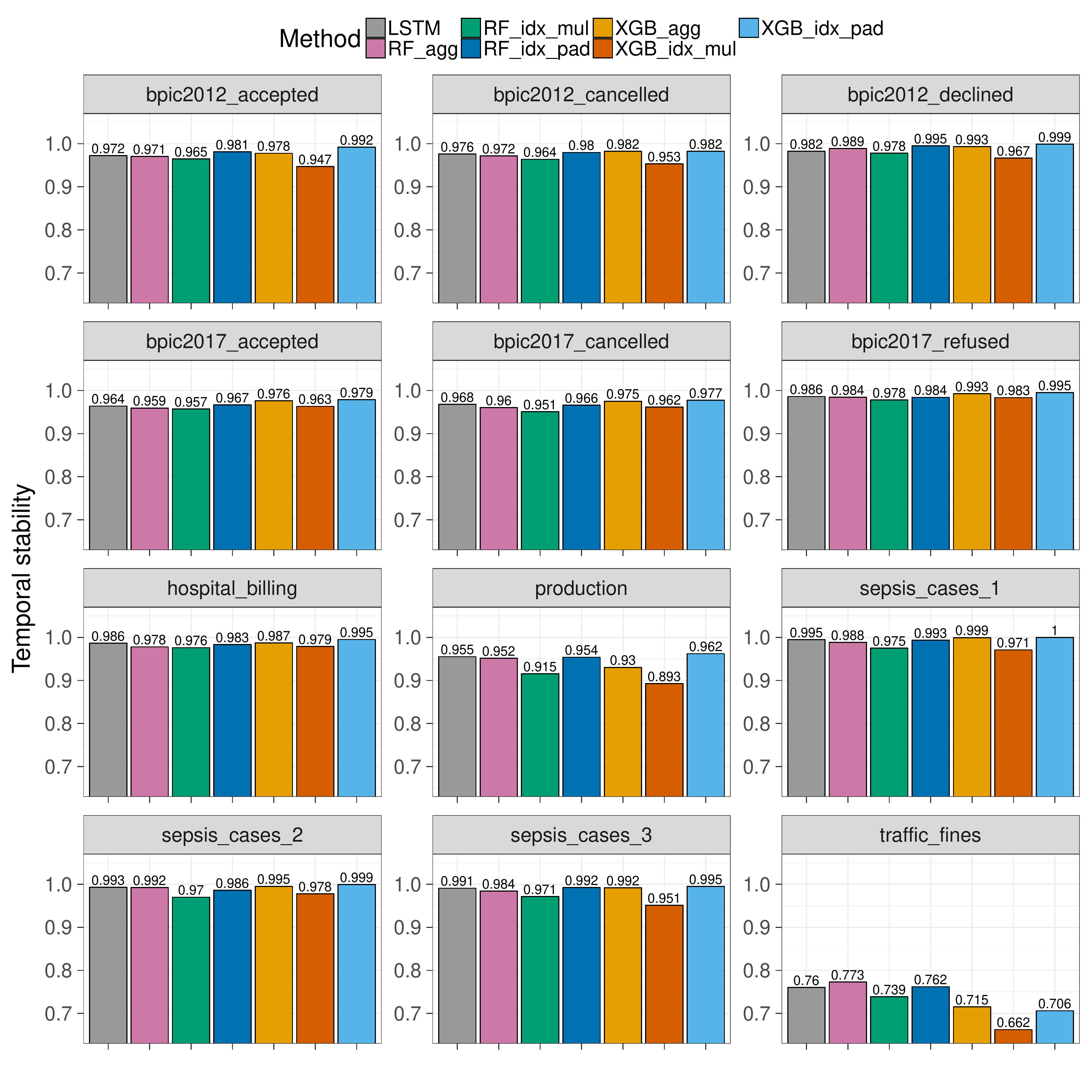}
\caption{Temporal stability.}
\label{fig:stability_mean_abs_calibrated}
\end{figure}

The temporal stability is plotted in \figurename~\ref{fig:stability_mean_abs_calibrated}.
In 11 out of 12 datasets, the highest stability is achieved by \emph{XGB\_idx\_pad}, usually followed by \emph{XGB\_agg} and then either \emph{LSTM} or \emph{RF\_idx\_pad}. In general, RF achieves slightly lower stability than its XGBoost counterparts. The multiclassifier approaches always have lower temporal stability than single classifiers, which is not surprising. Namely, as the RF and XGBoost classifiers do not consider the temporal relations between the input features and, instead, assume them to be i.i.d., the variance between classifiers built for prefixes of length $l$ and $l+1$ can be very high and, thus, the predictions made for two successive prefixes can be completely uncorrelated. This discussion answers RQ1.

\subsubsection{Increasing the inter-run stability during validation}

Tables~\ref{table:optim_auc} and \ref{table:optim_stability} present the overall AUC (weighted average over all prefix lengths) and the temporal stability for the single classifier with aggregation encoding with RF and XGBoost using three hyperparameter optimization approaches: 1) validation based on AUC over a single run with each parameter setting (\emph{RF}, \emph{XGB}), 2) validation based on average AUC over 5 runs with each parameter setting (\emph{RF\_5}, \emph{XGB\_5}), and 3) validation based on a combined measure of mean AUC and inter-run stability over 5 runs with each parameter setting (\emph{RF\_5\_S}, \emph{XGB\_5\_S}).

The results show that selecting the best parameters according to AUC over 5 runs usually (in 7 out of 12 cases) increases the AUC on the test set as compared to selecting them based on a single run, while the temporal stability is increased almost always (the only exceptions are \emph{traffic\_fines} and \emph{sepsis\_cases\_3}). Optimizing the combined metric over 5 runs further improves the temporal stability, but achieves slightly less consistent improvement in AUC. The validation over 5 runs increases the temporal stability also for XGBoost. In fact, the highest temporal stability is achieved by either \emph{XGB\_5} or \emph{XGB\_5\_S} in the majority of the datasets as can be seen in Table~\ref{table:optim_stability}. The AUC in the case of XGBoost remains at the same level or even decreases as compared to validating over a single run. The best AUC is often achieved by \emph{RF\_5} or \emph{RF\_5\_S} (Table~\ref{table:optim_auc}).

To answer RQ2, we found that validating over 5 runs instead of a single run, in general, results in improvement of AUC and/or temporal stability. However, the improvements are rather small in value and come at the expense of running 5 times more experiments during the hyperparameter optimization phase.

\begin{table}[t] \centering
  \caption{Effects of maximizing inter-run stability and accuracy on AUC.}
  \label{table:optim_auc}
{\scriptsize
\begin{tabular}{@{\extracolsep{5pt}} ccccccc}
\\[-1.8ex]\hline
\hline \\[-1.8ex]
dataset & RF & RF\_5 & RF\_5\_S & XGB & XGB\_5 & XGB\_5\_S \\
\hline \\[-1.8ex]
bpic2012\_accepted & $\mathbf{0.690}$ & $\mathbf{0.690}$ & $0.674$ & $0.680$ & $0.677$ & $0.677$ \\ 
bpic2012\_cancelled & $\mathbf{0.700}$ & $0.691$ & $0.688$ & $0.697$ & $0.690$ & $0.695$ \\ 
bpic2012\_declined & $\mathbf{0.610}$ & $0.609$ & $0.609$ & $0.605$ & $0.599$ & $0.603$ \\ 
bpic2017\_accepted & $0.834$ & $\mathbf{0.843}$ & $0.839$ & $0.834$ & $0.841$ & $0.831$ \\ 
bpic2017\_cancelled & $0.803$ & $\mathbf{0.813}$ & $0.812$ & $0.810$ & $0.811$ & $0.812$ \\ 
bpic2017\_refused & $0.805$ & $0.816$ & $\mathbf{0.820}$ & $0.802$ & $0.810$ & $0.801$ \\ 
hospital\_billing & $0.671$ & $0.662$ & $0.665$ & $\mathbf{0.731}$ & $0.727$ & $0.724$ \\ 
production & $\mathbf{0.707}$ & $0.540$ & $0.540$ & $0.565$ & $0.563$ & $0.563$ \\ 
sepsis\_cases\_1 & $0.611$ & $\mathbf{0.638}$ & $\mathbf{0.638}$ & $0.512$ & $0.490$ & $0.490$ \\ 
sepsis\_cases\_2 & $0.750$ & $\mathbf{0.781}$ & $0.763$ & $0.761$ & $0.742$ & $0.683$ \\ 
sepsis\_cases\_3 & $0.693$ & $\mathbf{0.747}$ & $\mathbf{0.747}$ & $0.738$ & $0.712$ & $0.712$ \\ 
traffic\_fines & $0.667$ & $\mathbf{0.681}$ & $\mathbf{0.681}$ & $0.661$ & $0.661$ & $0.660$ \\ 
\hline \\[-1.8ex]
\end{tabular}
}
\end{table}

\begin{table}[t] \centering
  \caption{Effects of maximizing inter-run stability and accuracy on temporal stability.}
  \label{table:optim_stability}
{\scriptsize
\begin{tabular}{@{\extracolsep{5pt}} ccccccc}
\\[-1.8ex]\hline
\hline \\[-1.8ex]
dataset & RF & RF\_5 & RF\_5\_S & XGB & XGB\_5 & XGB\_5\_S \\
\hline \\[-1.8ex]
bpic2012\_accepted & $0.971$ & $0.970$ & $0.974$ & $0.978$ & $0.988$ & $\mathbf{0.994}$ \\ 
bpic2012\_cancelled & $0.972$ & $0.970$ & $0.977$ & $0.982$ & $0.991$ & $\mathbf{0.996}$ \\ 
bpic2012\_declined & $0.989$ & $0.988$ & $0.988$ & $0.993$ & $0.993$ & $\mathbf{0.996}$ \\ 
bpic2017\_accepted & $0.959$ & $0.974$ & $0.975$ & $0.976$ & $\mathbf{0.988}$ & $0.977$ \\ 
bpic2017\_cancelled & $0.960$ & $0.973$ & $0.974$ & $0.975$ & $\mathbf{0.989}$ & $0.976$ \\ 
bpic2017\_refused & $0.984$ & $0.991$ & $0.992$ & $0.993$ & $\mathbf{0.998}$ & $0.992$ \\ 
hospital\_billing & $0.978$ & $0.976$ & $0.977$ & $0.987$ & $0.980$ & $\mathbf{0.981}$ \\ 
production & $0.952$ & $0.939$ & $0.939$ & $0.930$ & $\mathbf{0.999}$ & $\mathbf{0.999}$ \\ 
sepsis\_cases\_1 & $0.988$ & $0.993$ & $0.993$ & $0.999$ & $\mathbf{1.000}$ & $\mathbf{1.000}$ \\ 
sepsis\_cases\_2 & $0.992$ & $0.990$ & $0.993$ & $0.995$ & $0.994$ & $\mathbf{1.000}$ \\ 
sepsis\_cases\_3 & $0.984$ & $0.982$ & $0.982$ & $\mathbf{0.992}$ & $0.987$ & $0.987$ \\ 
traffic\_fines & $\mathbf{0.773}$ & $0.769$ & $0.769$ & $0.715$ & $0.697$ & $0.702$ \\
\hline \\[-1.8ex]
\end{tabular}}
\end{table}

\subsubsection{Decreasing the intra-case prediction volatility during prediction}
\label{sec:smoothing_results}

\begin{figure}[t]
\centering
\includegraphics[width=1\textwidth]{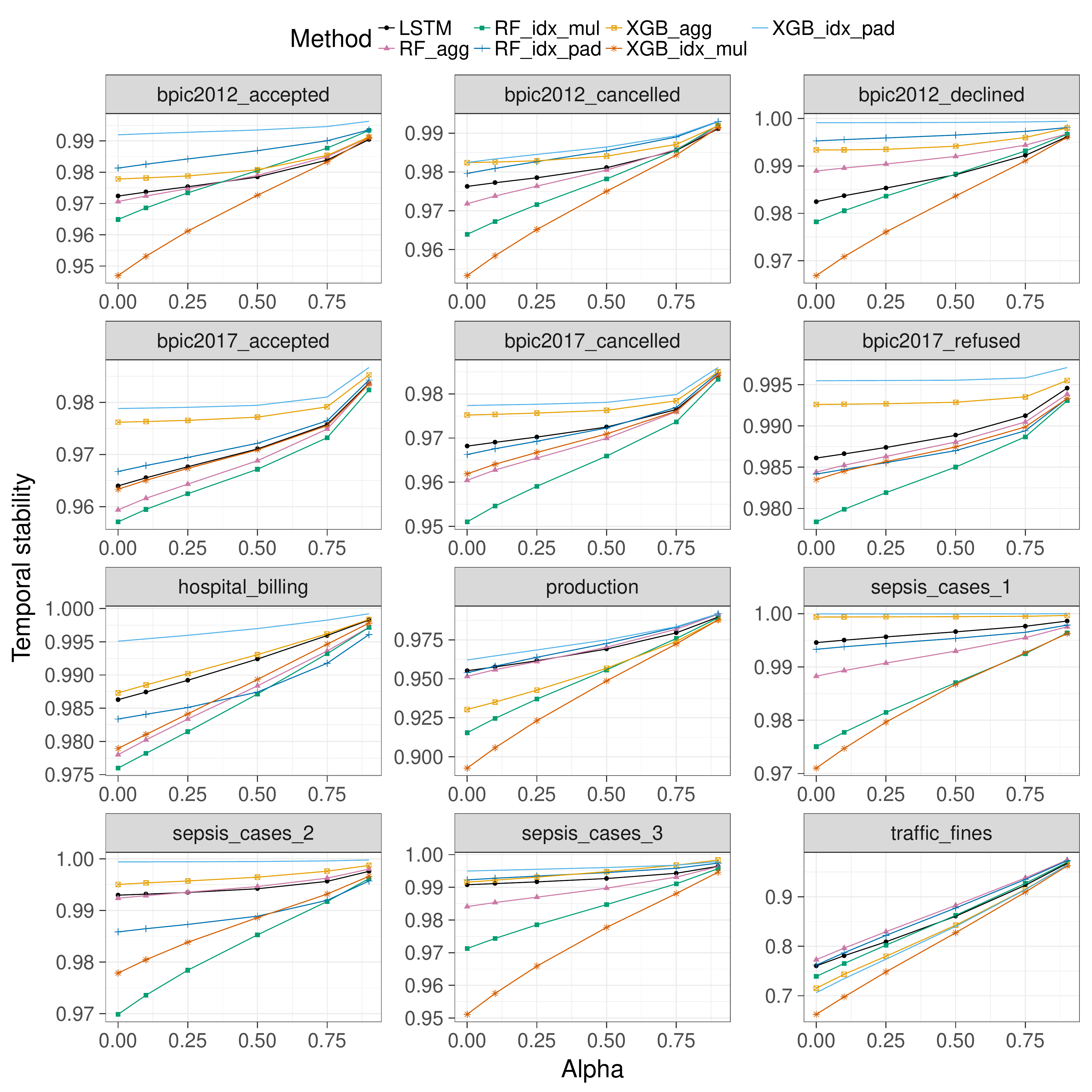}
\caption{Temporal stability across different levels of smoothing.}
\label{fig:stability_all}
\end{figure}

\figurename~\ref{fig:stability_all} shows that decreasing the prediction volatility via exponential smoothing consistently improves the temporal stability. The larger the smoothing parameter $\alpha$, the larger the increase in temporal stability. The methods that benefit the most from smoothing are multiclassifiers (\emph{RF\_idx\_mul} and \emph{XGB\_idx\_mul}). Being initially less stable, smoothing helps these methods to achieve a similar level of temporal stability as the other methods. In some cases, the multiclassifiers even overtake the other methods on large $\alpha$ (see \emph{bpic2012} variants). Also, \emph{RF\_agg} gains relatively more from smoothing than its XGBoost counterpart and \emph{LSTM}. For instance, see \emph{bpic2012} variants or \emph{production}, where \emph{RF\_agg} bypasses either \emph{LSTM} or \emph{XGB\_agg}.

\begin{figure}[t]
\centering
\includegraphics[width=1\textwidth]{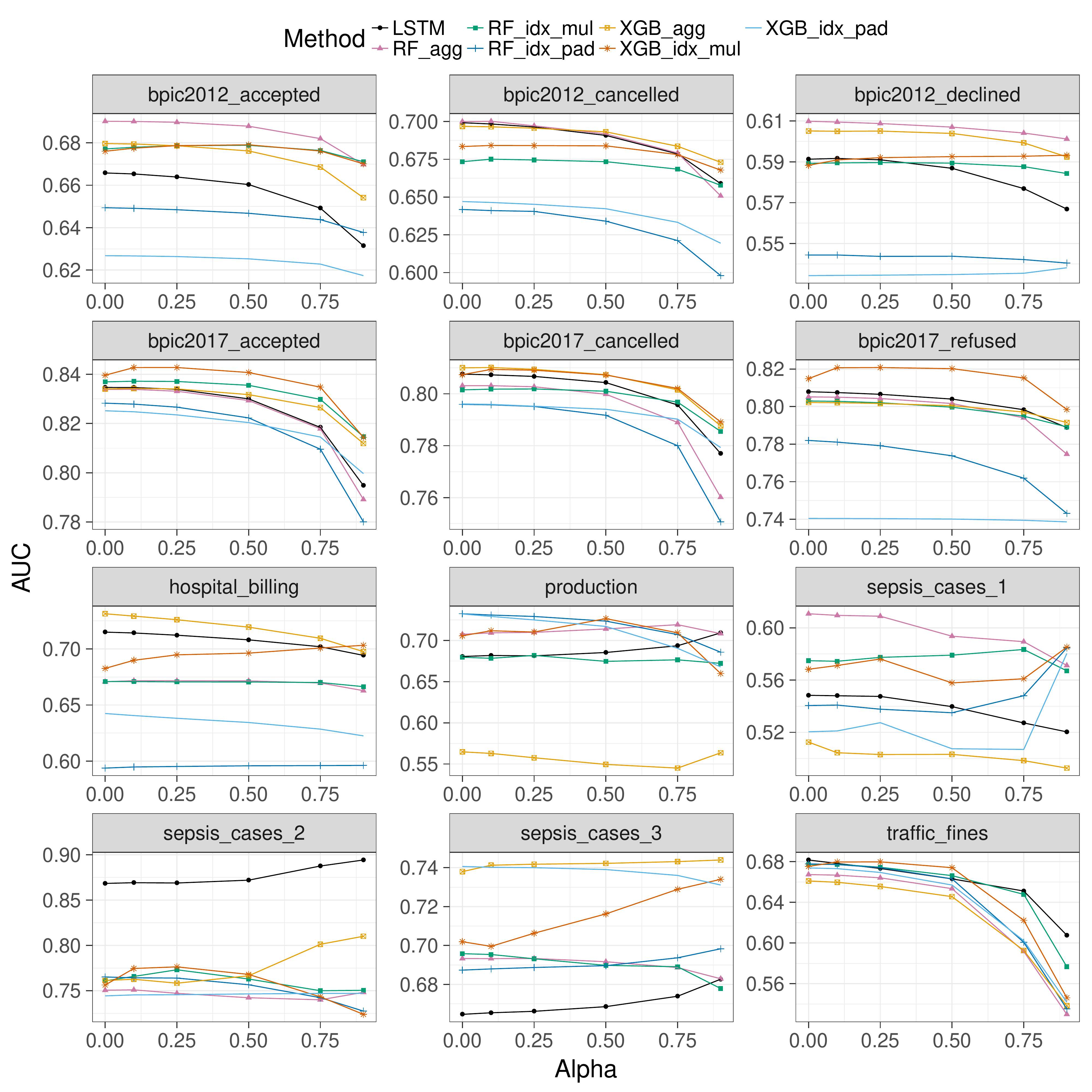}
\caption{Overall prediction accuracy across different levels of smoothing.}
\label{fig:aucs_all}
\end{figure}

In \figurename~\ref{fig:aucs_all}, the overall AUC is plotted against the $\alpha$ parameter. We observe that in most cases smoothing decreases the AUC. The reason for this is that as the smoothed estimate is cautious about the most recent prediction, the true signal in the data occurs after a lag. However, the AUC does not always decrease with smoothing. For smaller logs (\emph{production} and \emph{sepsis\_cases} variants), the AUC remains almost unchanged by smoothing or even increases. Also in the larger logs, a small amount of smoothing can help to increase the AUC (e.g., see \emph{XGB\_idx\_mul} in \emph{bpic2017\_refused}). The methods that benefit the most from smoothing are again the multiclassifiers. While not the most accurate methods before postprocessing, they often overtake the other methods with high levels of smoothing.

\begin{figure}[t]
\centering
\includegraphics[width=1\textwidth]{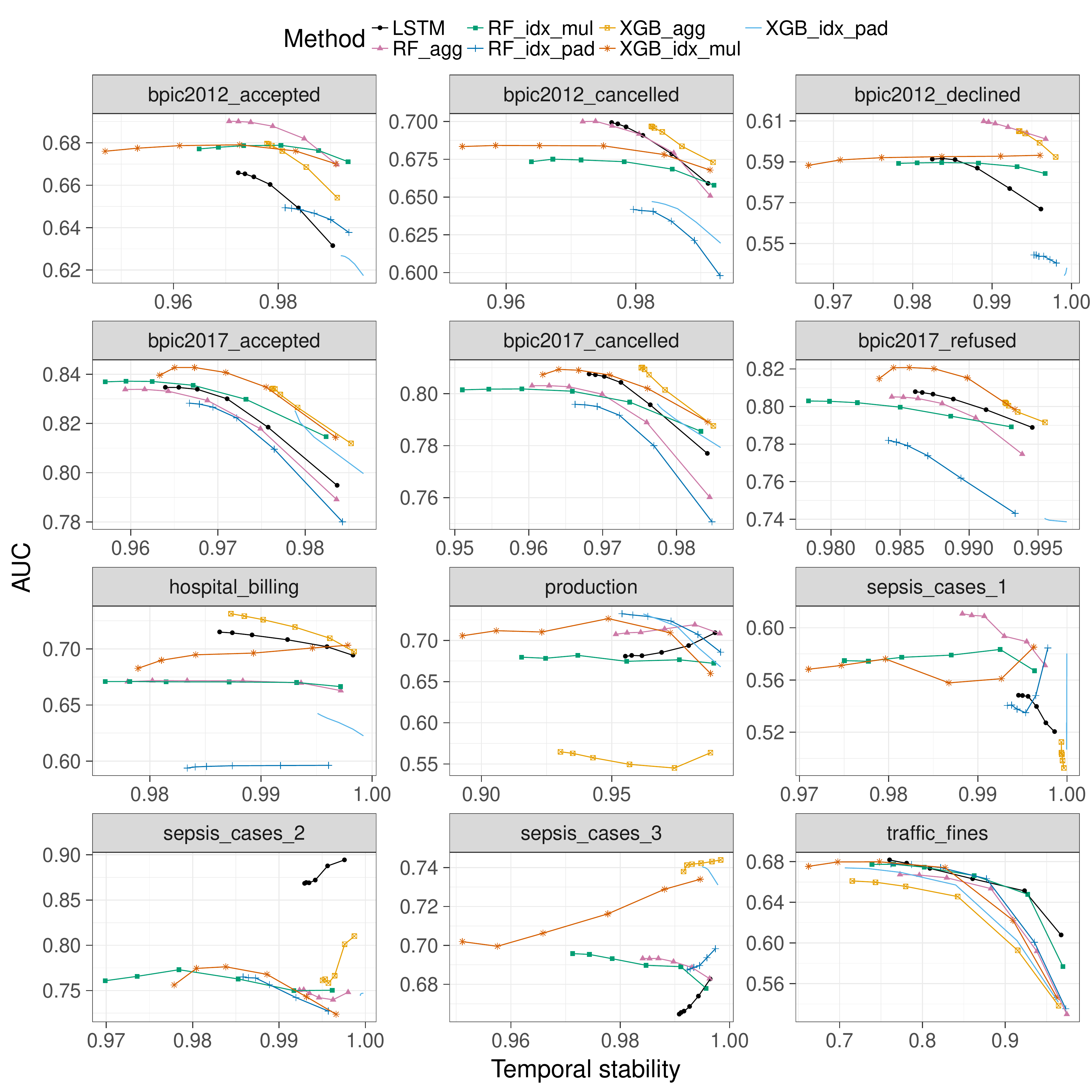}
\caption{Temporal stability vs. prediction accuracy.}
\label{fig:stability_vs_auc}
\end{figure}

To further understand the relationship between AUC and temporal stability, let us look at \figurename~\ref{fig:stability_vs_auc}, where these two metrics are plotted against each other (each dot corresponds to AUC and temporal stability obtained via smoothing with a particular value of $\alpha$). We see that \emph{RF\_idx\_mul} and \emph{XGB\_idx\_mul} change considerably in the direction from left to right, indicating that they are initially unstable but improve substantially with smoothing. At the same time, their change in the up-down direction is small, meaning that the AUC is not affected much. 
 The least affected by smoothing is the \emph{XGB\_idx\_pad} method. For instance, in \emph{bpic2012\_declined} and \emph{sepsis\_cases\_2} both the accuracy and the temporal stability remain almost constant. We also observe that, although the \emph{LSTM} method in the smaller logs is initially stable and does not gain in stability when smoothing, it does benefit in terms of AUC in the cases of \emph{production}, \emph{sepsis\_cases\_2}, and \emph{sepsis\_cases\_3}.
The \emph{XGB\_agg} method often appears in the top right corner, dominating the other techniques in terms of both accuracy and stability (see, for instance, \emph{bpic2012\_cancelled}, \emph{bpic2017\_cancelled}, \emph{hospital\_billing}, and \emph{sepsis\_cases\_3}).

To answer RQ3, exponential smoothing helps to increase the temporal stability, but usually at the expense of lower accuracy. Exceptions are \emph{RF\_idx\_mul} and \emph{XGB\_idx\_mul}, where smoothing often increases both temporal stability and AUC.

\section{Conclusion and Future Work}
\label{sec:conclusion}

We introduced the notion of temporal stability for predictive process monitoring. Temporal stability characterizes how much successive prediction scores obtained for the same case (sequence of events) differ from each other. For a temporally stable classifier, such successive prediction scores are similar to each other, resulting in a smooth time series, while in case of an unstable classifier, the resulting time series is volatile.
We evaluated the temporal stability of 7 existing predictive process monitoring methods, including single and multiclassifiers using RF, XGBoost, and LSTM. The experiments were done on 12 prediction tasks formulated on 6 real-life publicly available datasets. We found that the highest temporal stability was achieved by a single classifier approach with XGBoost (using either aggregation or index-based encoding), followed by LSTM.

We investigated the effects of hyperparameter optimization on temporal stability. We compared the final classifiers constructed after selecting the best parameters based on 1) AUC over a single run for each parameter setting, 2) AUC over 5 runs for each setting, 3) combined AUC and inter-run stability over 5 runs for each setting. The results show that choosing the parameters based on 5 runs can increase both AUC and temporal stability. However, the improvement is small and is subject to the trade-off of 5 times more computations during validation.

Finally, we explored how exponential smoothing affects the AUC and temporal stability.
We concluded that smoothing can be a reasonable approach for adjusting the predictions in applications where temporal stability is important at the expense of achieving slightly smaller AUC. Moreover, we observed that the multiclassifiers benefit the most from smoothing, in some cases even increasing both the temporal stability and the AUC at the same time. Therefore, when high temporal stability is required, it may be reasonable to use a multiclassifier approach with smoothing, achieving stable results with little or no loss in accuracy.

As future work, we plan to develop more robust notions of temporal stability that would still require most of the successive differences in predictions to be small, but not penalize the classifier for changing the prediction when an event with a relevant signal arrives. We will examine if the works on early sequence classification could be helpful in developing an adaptive smoothing method that decreases volatility on subsequences without suppressing the relevant signal.
Furthermore, the notion of temporal stability could be extended to other prediction tasks, such as multi-class predictions and regression. For instance, temporal stability could also be investigated in the context of predicting the remaining time of an ongoing case. While several methods have been developed with the goal of providing accurate remaining time estimations, using, e.g., non-parametric regression~\cite{van2008cycle}, support vector regression~\cite{polato2014data}, or LSTM neural networks~\cite{TaxVRD17}, none of these works has considered the stability of the predictions.
Another avenue for future work is to incorporate the notion of stability into the training phase of the classifiers. For instance, in case of neural networks this could be achieved by adjusting the loss function to take into account both the accuracy and the stability of the predictions.

\smallskip\noindent
\textbf{Acknowledgments}. This research was partly funded by the Estonian Research Council (grant IUT20-55).

\bibliographystyle{spbasic}      
\bibliography{bibliography}   

\clearpage

\appendix
\section*{Appendix}
\label{appendix}


\begin{figure}[hbtp]
\centering
\includegraphics[width=0.86\textwidth]{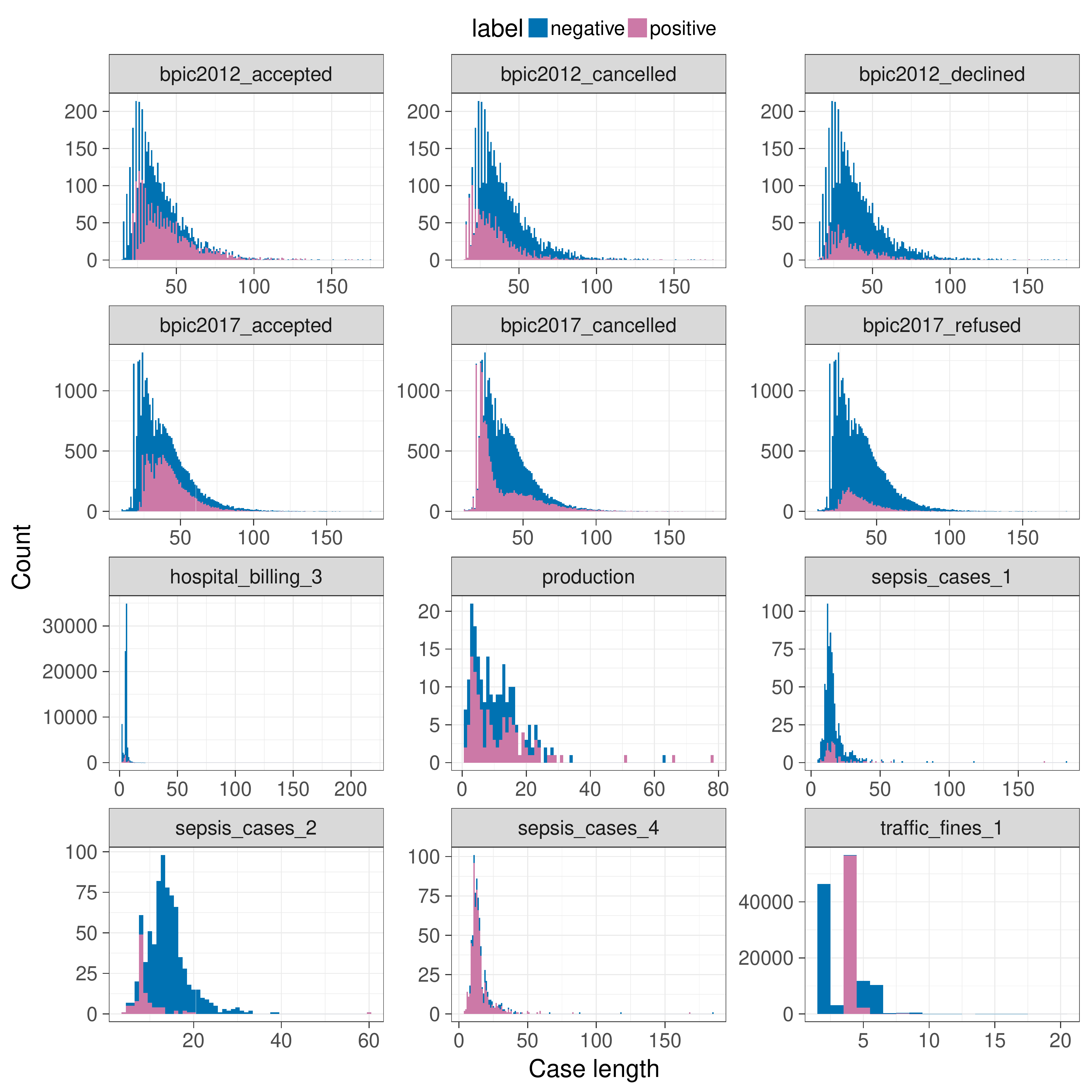}
\caption{Case length histograms for positive and negative classes.}
\label{fig:case_length_hist}
\end{figure}

\begin{figure}[hbtp]
\centering
\includegraphics[width=1\textwidth]{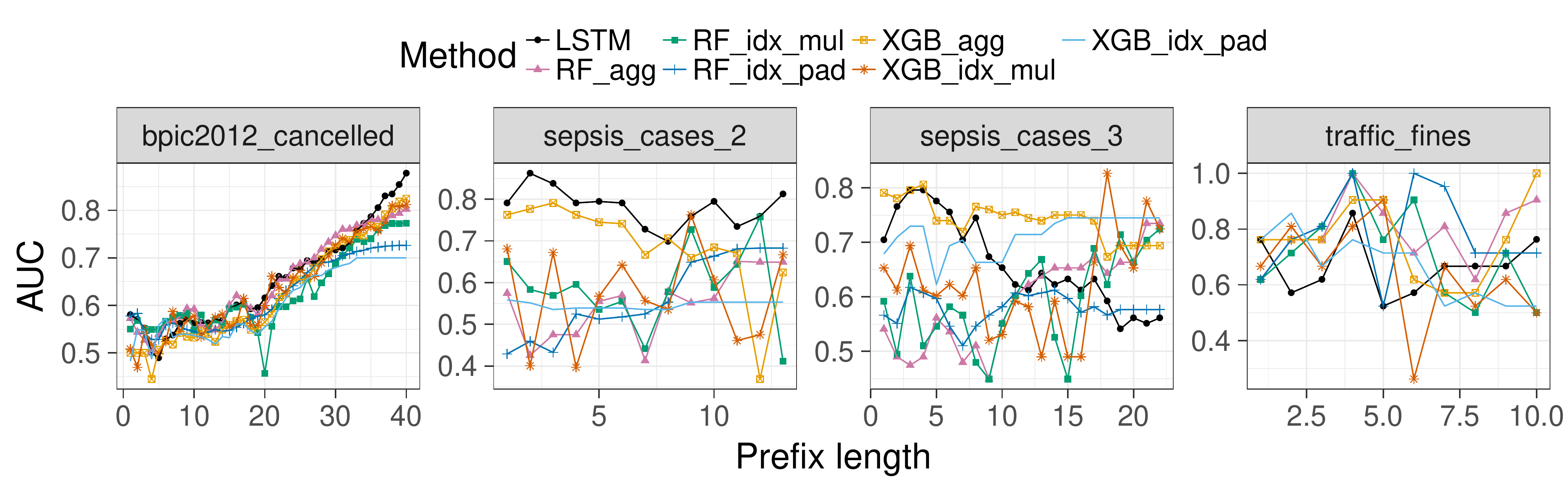}
\caption{Prediction accuracy on long cases only.}
\label{fig:aucs_calibrated_max_prefix}
\end{figure}


\begin{table}[hbtp]
\caption{Hyperparameters and distributions used in optimization via random search.}
\label{table:hyperparameters}
\begin{center}
\begin{tabular}{@{}llll@{}}
\toprule
Classifier & Parameter & Distribution & Values \\
\midrule
\multirow{2}{*}{RF} & \# estimators (\emph{n\_est}) & Uniform integer & $x \in [150, 1000]$ \\
& Max features (\emph{mf}) & Log-uniform & $x \in [0.01, 0.9]$ \\
\midrule
\multirow{6}{*}{XGBoost} & \# estimators (\emph{n\_est}) & Uniform integer & $x \in [150, 1000]$ \\
& Learning rate (\emph{lr}) & Uniform & $x \in [0.01, 0.07]$ \\
& Subsample (\emph{subs}) & Uniform & $x \in [0.5, 1]$ \\
& Max tree depth (\emph{md}) & Uniform integer & $x \in [3, 9]$ \\
& Colsample bytree (\emph{cb}) & Uniform & $x \in [0.5, 1]$ \\
& Min child weight (\emph{mcw}) & Uniform integer & $x \in [1, 3]$ \\
\midrule
\multirow{6}{*}{LSTM} & \# hidden layers (\emph{n\_lay}) & Categorical & $x \in \{1, 2, 3\}$ \\
& \# units in hidden layer (\emph{n\_hid}) & Log-uniform integer & $x \in [10, 150]$ \\
& Initial learning rate (\emph{lr}) & Log-uniform & $x \in [0.000001, 0.0001]$ \\
& Batch size (\emph{batch}) & Categorical & $x \in \{8, 16, 32, 64\}$ \\
& Dropout (\emph{drop}) & Uniform & $x \in [0, 0.3]$ \\
& Optimizer (\emph{opt}) & Categorical & $x \in \{RMSProp, NAdam\}$ \\
\bottomrule
\end{tabular}
\end{center}
\end{table}

\begin{table}[t]
\caption{Optimized hyperparameters (RF).}
\label{table:optimal_params_rf}
\begin{center}
\resizebox{1\textwidth}{!}{
	\begin{tabular}{@{}lcccccccccccc@{}}
	\toprule
 & \multicolumn{2}{c}{RF\_agg}  & \multicolumn{2}{c}{RF\_idx\_pad} & \multicolumn{2}{c}{RF\_idx\_mul}  & \multicolumn{2}{c}{RF\_idx\_mul} & \multicolumn{2}{c}{RF\_idx\_mul}  & \multicolumn{2}{c}{RF\_idx\_mul} \\  
 & \multicolumn{2}{c}{}  & \multicolumn{2}{c}{} & \multicolumn{2}{c}{prefix=1}  & \multicolumn{2}{c}{prefix=5} & \multicolumn{2}{c}{prefix=10}  & \multicolumn{2}{c}{prefix=20} \\  
dataset & n\_est & mf & n\_est & mf & n\_est & mf & n\_est & mf & n\_est & mf & n\_est & mf \\ \midrule
production & 769 & 0.02 & 769 & 0.02 & 468 & 0.35 & 556 & 0.73 & 944 & 0.06 & 833 & 0.86 \\
sepsis\_cases\_1 & 873 & 0.02 & 537 & 0.13 & 844 & 0.03 & 840 & 0.79 & 408 & 0.25 & 739 & 0.52 \\
sepsis\_cases\_2 & 313 & 0.25 & 873 & 0.02 & 990 & 0.6 & 567 & 0.29 & 316 & 0.07 & - & - \\
sepsis\_cases\_3 & 537 & 0.13 & 313 & 0.25 & 269 & 0.6 & 623 & 0.25 & 886 & 0.32 & 614 & 0.02 \\
traffic\_fines & 847 & 0.27 & 847 & 0.27 & 593 & 0.62 & 912 & 0.38 & 206 & 0.22 & - & - \\
bpic2012\_accepted & 801 & 0.07 & 958 & 0.01 & 474 & 0.24 & 273 & 0.85 & 287 & 0.03 & 273 & 0.85 \\
bpic2012\_cancelled & 324 & 0.06 & 958 & 0.01 & 273 & 0.85 & 751 & 0.37 & 673 & 0.03 & 635 & 0.75 \\
bpic2012\_declined & 801 & 0.07 & 675 & 0.35 & 635 & 0.75 & 635 & 0.75 & 273 & 0.85 & 287 & 0.03 \\
bpic2017\_accepted & 511 & 0.14 & 828 & 0.14 & 445 & 0.17 & 609 & 0.33 & 805 & 0.41 & 685 & 0.07 \\
bpic2017\_refused & 511 & 0.14 & 828 & 0.14 & 863 & 0.07 & 560 & 0.24 & 537 & 0.4 & 537 & 0.4 \\
bpic2017\_cancelled & 511 & 0.14 & 828 & 0.14 & 805 & 0.41 & 152 & 0.33 & 362 & 0.44 & 782 & 0.43 \\
hospital\_billing & 549 & 0.13 & 277 & 0.22 & 969 & 0.01 & 969 & 0.01 & - & - & - & - \\
	\bottomrule
	\end{tabular}}
\end{center}
\end{table}

\begin{table}[t]
\caption{Optimized hyperparameters for single classifiers (XGBoost).}
\label{table:optimal_params_xgboost_single}
\begin{center}
\resizebox{1\textwidth}{!}{
	\begin{tabular}{@{}lcccccccccccc@{}}
	\toprule
 & \multicolumn{6}{c}{XGB\_agg}  & \multicolumn{6}{c}{XGB\_idx\_pad}  \\  
dataset & n\_est & lr & subs & md & cb & mcw & n\_est & lr & subs & md & cb & mcw \\ \midrule
production & 224 & 0.01 & 0.53 & 5 & 0.95 & 1 & 699 & 0.07 & 0.77 & 8 & 0.63 & 2 \\
sepsis\_cases\_1 & 355 & 0.02 & 0.59 & 3 & 0.91 & 2 & 399 & 0.06 & 0.68 & 8 & 0.87 & 2 \\
sepsis\_cases\_2 & 971 & 0.04 & 0.73 & 8 & 0.73 & 2 & 476 & 0.04 & 0.52 & 4 & 0.72 & 1 \\
sepsis\_cases\_3 & 355 & 0.02 & 0.59 & 3 & 0.91 & 2 & 918 & 0.02 & 0.78 & 8 & 0.97 & 1 \\
traffic\_fines & 773 & 0.04 & 0.75 & 7 & 0.71 & 2 & 773 & 0.04 & 0.75 & 7 & 0.71 & 2 \\
bpic2012\_accepted & 156 & 0.01 & 0.78 & 8 & 0.61 & 1 & 710 & 0.01 & 0.51 & 7 & 0.78 & 1 \\
bpic2012\_cancelled & 445 & 0.03 & 0.9 & 5 & 0.61 & 1 & 291 & 0.05 & 0.85 & 7 & 0.79 & 2 \\
bpic2012\_declined & 363 & 0.05 & 0.8 & 3 & 0.65 & 2 & 363 & 0.05 & 0.8 & 3 & 0.65 & 2 \\
bpic2017\_accepted & 215 & 0.03 & 0.75 & 4 & 0.68 & 1 & 830 & 0.01 & 0.62 & 5 & 0.84 & 2 \\
bpic2017\_refused & 215 & 0.03 & 0.75 & 4 & 0.68 & 1 & 830 & 0.01 & 0.62 & 5 & 0.84 & 2 \\
bpic2017\_cancelled & 187 & 0.04 & 0.76 & 4 & 0.79 & 1 & 830 & 0.01 & 0.62 & 5 & 0.84 & 2 \\
hospital\_billing & 215 & 0.03 & 0.75 & 4 & 0.68 & 1 & 735 & 0.06 & 0.71 & 3 & 0.54 & 1 \\
	\bottomrule
	\end{tabular}}
\end{center}
\end{table}

\begin{table}[t]
\caption{Optimized hyperparameters for multiclassifiers (XGBoost).}
\label{table:optimal_params_xgboost_multi}
\begin{center}
\resizebox{0.95\textwidth}{!}{
	\begin{tabular}{@{}lcccccccccccc@{}}
	\toprule
 & \multicolumn{6}{c}{XGB\_idx\_mul, prefix=1}  & \multicolumn{6}{c}{XGB\_idx\_mul, prefix=5}  \\  
dataset & n\_est & lr & subs & md & cb & mcw & n\_est & lr & subs & md & cb & mcw \\ \midrule
production & 228 & 0.03 & 0.63 & 3 & 0.98 & 1 & 436 & 0.03 & 0.98 & 3 & 0.85 & 2 \\
sepsis\_cases\_1 & 918 & 0.02 & 0.78 & 8 & 0.97 & 1 & 187 & 0.06 & 0.82 & 7 & 0.85 & 2 \\
sepsis\_cases\_2 & 971 & 0.04 & 0.73 & 8 & 0.73 & 2 & 187 & 0.06 & 0.82 & 7 & 0.85 & 2 \\
sepsis\_cases\_3 & 764 & 0.02 & 0.93 & 4 & 0.52 & 1 & 712 & 0.01 & 0.89 & 5 & 0.92 & 1 \\
traffic\_fines & 977 & 0.02 & 0.52 & 7 & 0.82 & 1 & 615 & 0.03 & 0.73 & 6 & 0.59 & 2 \\
bpic2012\_accepted & 394 & 0.06 & 0.98 & 8 & 0.5 & 2 & 291 & 0.05 & 0.85 & 7 & 0.79 & 2 \\
bpic2012\_cancelled & 819 & 0.06 & 0.99 & 4 & 0.88 & 1 & 190 & 0.02 & 0.83 & 5 & 1.0 & 1 \\
bpic2012\_declined & 363 & 0.05 & 0.8 & 3 & 0.65 & 2 & 445 & 0.03 & 0.9 & 5 & 0.61 & 1 \\
bpic2017\_accepted & 733 & 0.02 & 0.91 & 3 & 0.57 & 1 & 733 & 0.02 & 0.91 & 3 & 0.57 & 1 \\
bpic2017\_refused & 924 & 0.05 & 0.96 & 8 & 0.76 & 1 & 733 & 0.02 & 0.91 & 3 & 0.57 & 1 \\
bpic2017\_cancelled & 215 & 0.03 & 0.75 & 4 & 0.68 & 1 & 924 & 0.05 & 0.96 & 8 & 0.76 & 1 \\
hospital\_billing & 584 & 0.02 & 0.81 & 6 & 0.61 & 1 & 215 & 0.03 & 0.75 & 4 & 0.68 & 1 \\
\midrule
 & \multicolumn{6}{c}{XGB\_idx\_mul, prefix=10}  & \multicolumn{6}{c}{XGB\_idx\_mul, prefix=20}  \\  
dataset & n\_est & lr & subs & md & cb & mcw & n\_est & lr & subs & md & cb & mcw \\ \midrule
production & 921 & 0.03 & 0.52 & 6 & 0.82 & 1 & 909 & 0.06 & 0.86 & 6 & 0.89 & 2 \\
sepsis\_cases\_1 & 399 & 0.06 & 0.68 & 8 & 0.87 & 2 & 469 & 0.05 & 0.91 & 7 & 0.51 & 2 \\
sepsis\_cases\_2 & 971 & 0.04 & 0.73 & 8 & 0.73 & 2 & - & - & - & - & - & - \\
sepsis\_cases\_3 & 187 & 0.06 & 0.82 & 7 & 0.85 & 2 & 385 & 0.03 & 0.77 & 5 & 0.83 & 1 \\
traffic\_fines & 972 & 0.03 & 0.83 & 3 & 0.85 & 1 & - & - & - & - & - & - \\
bpic2012\_accepted & 720 & 0.01 & 0.89 & 7 & 0.93 & 1 & 445 & 0.03 & 0.9 & 5 & 0.61 & 1 \\
bpic2012\_cancelled & 526 & 0.05 & 0.79 & 7 & 0.88 & 2 & 190 & 0.02 & 0.83 & 5 & 1.0 & 1 \\
bpic2012\_declined & 156 & 0.01 & 0.78 & 8 & 0.61 & 1 & 445 & 0.03 & 0.9 & 5 & 0.61 & 1 \\
bpic2017\_accepted & 215 & 0.03 & 0.75 & 4 & 0.68 & 1 & 831 & 0.02 & 0.59 & 5 & 0.84 & 1 \\
bpic2017\_refused & 215 & 0.03 & 0.75 & 4 & 0.68 & 1 & 215 & 0.03 & 0.75 & 4 & 0.68 & 1 \\
bpic2017\_cancelled & 733 & 0.02 & 0.91 & 3 & 0.57 & 1 & 830 & 0.01 & 0.62 & 5 & 0.84 & 2 \\
hospital\_billing & - & - & - & - & - & - & - & - & - & - & - & - \\
	\bottomrule
	\end{tabular}}
\end{center}
\end{table}

\begin{table}[t]
\caption{Optimized hyperparameters (LSTM).}
\label{table:optimal_params_lstm}
\begin{center}
\resizebox{0.7\textwidth}{!}{
	\begin{tabular}{@{}lcccccc@{}}
	\toprule
 & \multicolumn{6}{c}{LSTM}  \\  
dataset & n\_lay & n\_hid & lr & batch & drop & opt \\ \midrule
production & 2 & 27 & 5e-05 & 16 & 0.05 & adam \\
sepsis\_cases\_1 & 2 & 27 & 3e-05 & 32 & 0.19 & nadam \\
sepsis\_cases\_2 & 1 & 80 & 4e-05 & 16 & 0.18 & nadam \\
sepsis\_cases\_3 & 2 & 46 & 4e-05 & 8 & 0.15 & nadam \\
traffic\_fines & 2 & 100 & 7e-05 & 16 & 0.27 & nadam \\
bpic2012\_accepted & 3 & 19 & 3e-05 & 8 & 0.18 & nadam \\
bpic2012\_cancelled & 2 & 21 & 2e-05 & 32 & 0.25 & nadam \\
bpic2012\_declined & 1 & 20 & 2e-05 & 32 & 0.02 & nadam \\
bpic2017\_accepted & 1 & 14 & 2e-05 & 8 & 0.03 & nadam \\
bpic2017\_refused & 1 & 10 & 4e-05 & 32 & 0.09 & nadam \\
bpic2017\_cancelled & 2 & 30 & 9e-05 & 64 & 0.11 & rmsprop \\
hospital\_billing & 3 & 144 & 5e-05 & 64 & 0.04 & rmsprop \\
	\bottomrule
	\end{tabular}}
\end{center}
\end{table}

\begin{table}[t]
\caption{Optimized hyperparameters (combined inter-run stability and AUC).}
\label{table:optimal_params_interrun}
\begin{center}
\resizebox{1\textwidth}{!}{
	\begin{tabular}{@{}lccccccccccccccccc@{}}
	\toprule
 & \multicolumn{2}{c}{RF\_5}  & \multicolumn{2}{c}{RF\_5\_S} & \multicolumn{6}{c}{XGB\_5}  & \multicolumn{6}{c}{XGB\_5\_S} \\
dataset & n\_est & mf & n\_est & mf & n\_est & lr & subs & md & cb & mcw & n\_est & lr & subs & md & cb & mcw \\ \midrule
production & 927 & 0.81 & 927 & 0.81 & 231 & 0.02 & 0.92 & 3 & 0.5 & 1 & 231 & 0.02 & 0.92 & 3 & 0.5 & 1 \\
sepsis\_cases\_1 & 858 & 0.1 & 858 & 0.1 & 586 & 0.01 & 0.76 & 3 & 0.97 & 1 & 586 & 0.01 & 0.76 & 3 & 0.97 & 1 \\
sepsis\_cases\_2 & 253 & 0.09 & 517 & 0.15 & 812 & 0.06 & 0.76 & 8 & 0.7 & 2 & 964 & 0.04 & 0.83 & 7 & 0.99 & 1 \\
sepsis\_cases\_3 & 764 & 0.11 & 764 & 0.11 & 154 & 0.03 & 0.57 & 7 & 0.56 & 2 & 154 & 0.03 & 0.57 & 7 & 0.56 & 2 \\
traffic\_fines & 957 & 0.24 & 957 & 0.24 & 424 & 0.05 & 0.71 & 8 & 0.76 & 2 & 669 & 0.02 & 0.99 & 4 & 0.67 & 1 \\
bpic2012\_accepted & 581 & 0.16 & 882 & 0.41 & 482 & 0.02 & 0.55 & 3 & 0.72 & 1 & 286 & 0.01 & 0.66 & 4 & 0.79 & 2 \\
bpic2012\_cancelled & 364 & 0.08 & 979 & 0.3 & 455 & 0.02 & 0.76 & 6 & 0.5 & 1 & 216 & 0.02 & 0.67 & 7 & 0.68 & 1 \\
bpic2012\_declined & 820 & 0.06 & 820 & 0.06 & 505 & 0.06 & 0.56 & 6 & 0.94 & 1 & 257 & 0.03 & 0.86 & 3 & 0.67 & 1 \\
bpic2017\_accepted & 169 & 0.41 & 359 & 0.46 & 284 & 0.02 & 0.78 & 7 & 0.57 & 1 & 499 & 0.03 & 0.69 & 3 & 0.97 & 1 \\
bpic2017\_refused & 346 & 0.24 & 410 & 0.54 & 933 & 0.01 & 0.65 & 6 & 0.87 & 1 & 325 & 0.06 & 0.92 & 3 & 0.8 & 1 \\
bpic2017\_cancelled & 301 & 0.26 & 300 & 0.41 & 161 & 0.01 & 0.89 & 6 & 0.51 & 2 & 276 & 0.03 & 0.79 & 4 & 0.72 & 1 \\
hospital\_billing & 900 & 0.07 & 969 & 0.08 & 921 & 0.02 & 0.75 & 7 & 0.99 & 2 & 730 & 0.01 & 0.63 & 8 & 0.97 & 2 \\
	\bottomrule
	\end{tabular}}
\end{center}
\end{table}


\begin{figure}[t]
\centering
\includegraphics[width=0.95\textwidth]{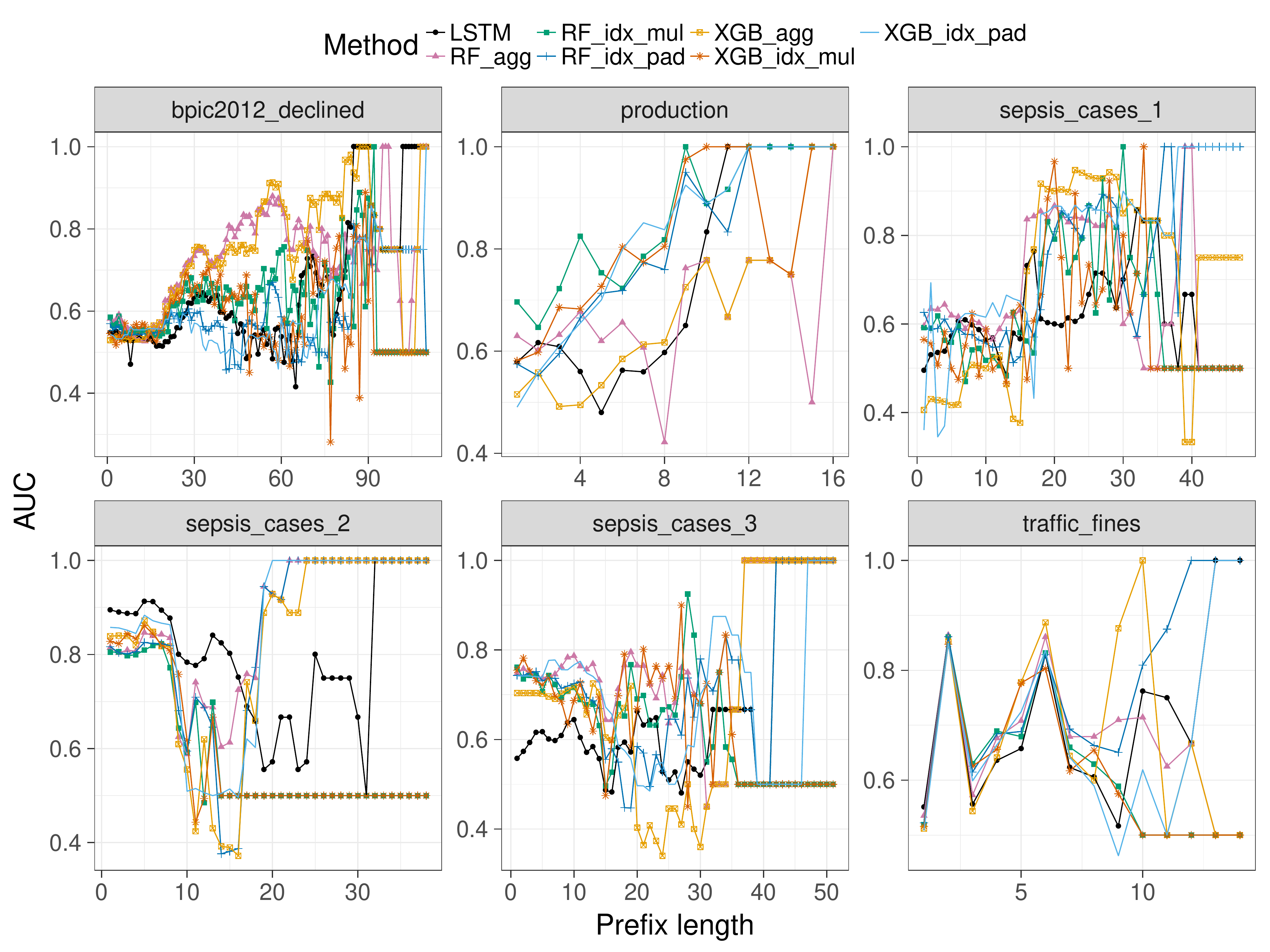}
\caption{Prediction accuracy on original (not truncated) traces.}
\label{fig:aucs_calibrated_not_truncated}
\end{figure}

\begin{figure}[t]
\centering
\includegraphics[width=1\textwidth]{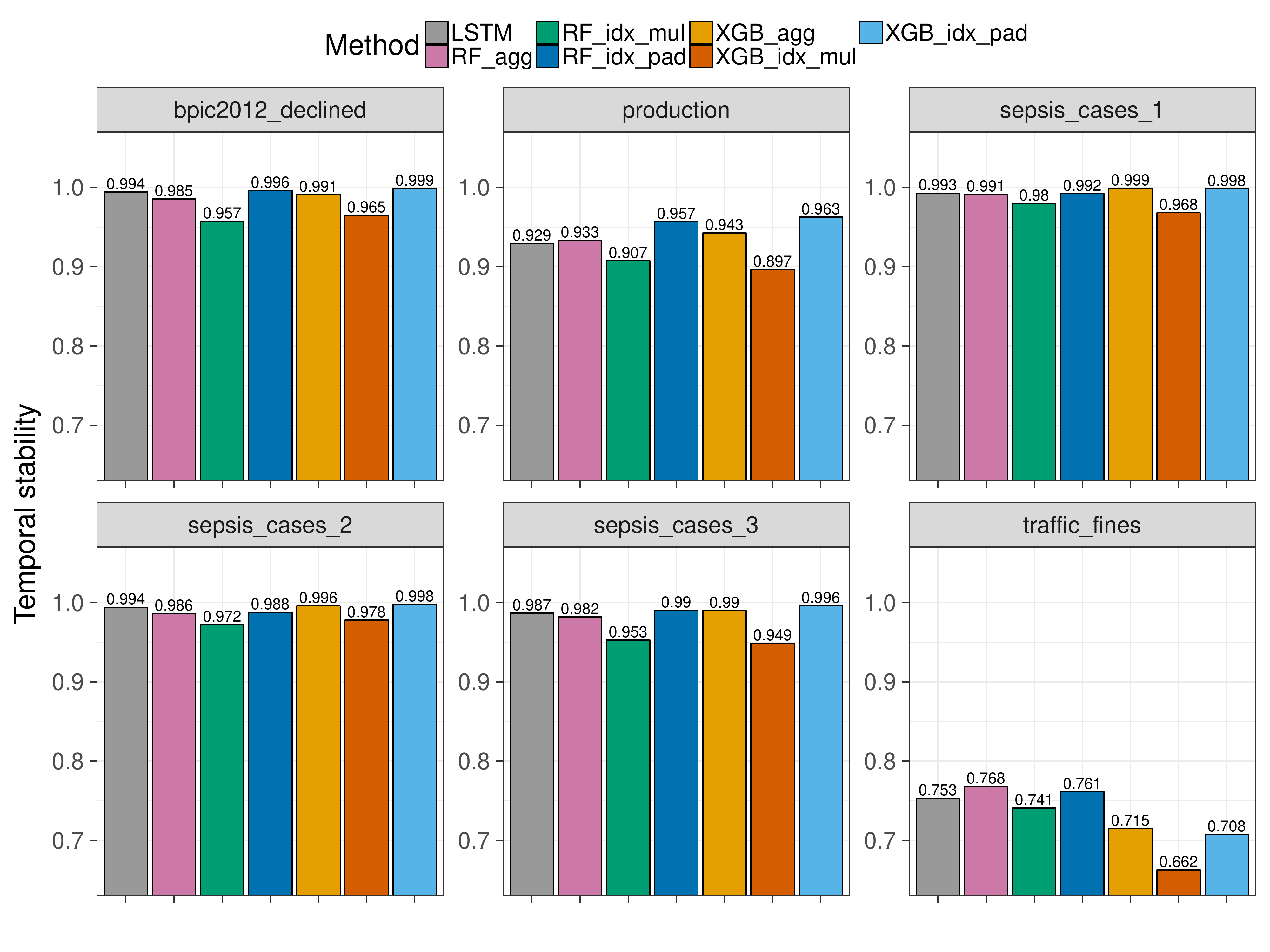}
\caption{Temporal stability on original (not truncated) traces.}
\label{fig:stability_mean_abs_calibrated_not_truncated}
\end{figure}

\end{document}